\documentclass[review]{elsarticle}
\usepackage{algorithm}
\usepackage{algorithmic}
\usepackage{makecell}
\usepackage{amsmath}
\usepackage{xcolor}
\usepackage{bm}
\usepackage{framed}
\usepackage{amssymb}
\usepackage{booktabs}
\usepackage{color}
\usepackage{multirow}
\usepackage{marvosym}
\usepackage{threeparttable}
\usepackage{subfigure}
\usepackage{graphicx}
\usepackage{epstopdf}
\usepackage{amsmath,amsthm}
 \usepackage{mathptmx}      
 \usepackage{marvosym}
 \usepackage{textcomp}

\usepackage{lineno,hyperref}

\begin{document}

\begin{frontmatter}

\title{GMC-PINNs: A new general Monte Carlo PINNs method for solving fractional partial differential equations on irregular domains}

\author[mymainaddress]{Shupeng Wang}
\ead{shupeng\_wang@brown.edu}

\author[mymainaddress]{George Em Karniadakis\corref{mycorrespondingauthor}}
\cortext[mycorrespondingauthor]{Corresponding author}
\ead{george\_karniadakis@brown.edu}


\address[mymainaddress]{Division of Applied Mathematics, Brown University, Providence, RI 02912, USA}

\begin{abstract}
Physics-Informed Neural Networks (PINNs) have been widely used for solving partial differential equations (PDEs) of different types, including fractional PDEs (fPDES) \cite{fpinn}. Herein, we propose a new general (quasi) Monte Carlo PINN for solving fPDEs on irregular domains. Specifically, instead of approximating fractional derivatives by Monte Carlo approximations of integrals as was done previously in \cite{lingguo}, we use a more general Monte Carlo approximation method to solve different fPDEs, which is valid for fractional differentiation under any definition. Moreover, based on the ensemble probability density function, the generated nodes are all located in denser regions near the target point where we perform the differentiation. This has an unexpected connection with known finite difference methods on non-equidistant or nested grids, and hence our method inherits their advantages. At the same time, the generated nodes exhibit a block-like dense distribution, leading to a good computational efficiency of this approach. We present the framework for using this algorithm and apply it to  several examples. Our results demonstrate the effectiveness of GMC-PINNs in dealing with irregular domain problems and show a higher computational efficiency compared to the original fPINN method. We also include comparisons with the Monte Carlo fPINN \cite{lingguo}. Finally, we use examples to demonstrate the effectiveness of the method in dealing with fuzzy boundary location problems, and then use the method to solve the coupled 3D fractional Bloch-Torrey equation defined in the ventricular domain of the human brain, and compare the results with classical numerical methods.
\end{abstract}


\end{frontmatter}

\section{Introduction}\label{sec1}
Fractional derivatives, as more general forms of derivatives, can describe physical systems with long-term memory and long-range interactions, so fractional partial differential equations (fPDEs) have been widely used in modeling systems with memory, spatial nonlocality, and power-law characteristics \cite{ks1993, george2016, colbrook2022, elia2013}. Since fractional integrals and fractional derivatives satisfy the commutation law only in special cases, no analytic solution to the definition of integral can be obtained even for linear problems. So many numerical methods for fPDEs have been developed \cite{tadj2006, zhangh2018, zhangh2019, george_book}, for example, \cite{zhangh_y2018} first developed the time-space spectral method and inverse problem for time-space fractional Fokker-Planck equation, which can enjoy high accuracy. Jia and Jiang \cite{JiaJiang2023} used exponential wave integrator method to study the improved uniform error bounds for long-time dynamics of the space fractional Klein-Gordon equation with weak nonlinearity. For high-dimensional time-space fPDEs on irregular domains, the main challenge is due to  L$\Acute{\mathrm{e}}$vy flights on how to choose a suitable adaptive mesh at the boundaries or use higher-order numerical techniques. Nevertheless, various classical methods have been generalized from low-dimensional regular domains to high-dimensional irregular domains \cite{bellman1971, lin2018, liu2015, pang2015, zay2015}. Specifically, the authors of \cite{jiang2017} proposed a new unstructured mesh Galerkin finite element method (FEM)  with Crank-Nicolson time-stepping for solving the 2D time-space fractional PDEs on an irregular convex domain. They showed that their method can achieve smaller errors than the structured mesh finite element method. For 3D problems, a class of RBF-based differential quadrature method was proposed in \cite{zhu2020} for the space fractional diffusion equation on 3D irregular domains. The radial basis function (RBF) was used as trial function, and the fractional derivatives were presented by weighted linear combinations. Numerical results showed that this method was convergent with desirable accuracy and compared favorably to finite difference method in terms of global errors. However, it is worth noting that due to the non-locality and singularity of fractional derivatives, high computational cost and high memory cost are required when using classical numerical methods to solve fPDEs. Although many fast algorithms \cite{ling2019, zeng2016, jia2021, zhang2023, bu2023} have been developed to improve computational efficiency for simple domains, it is still very important to develop more efficient numerical methods for high-dimensional fPDEs on irregular domains.

Recently, a new Monte Carlo method was proposed to approximate fractional-order differentiation \cite{podlubny2022,podlubny20222}. This method starts from the Gr\"{u}nwald-Letnikov definition of fractional derivative, constructs a probability distribution function that conforms to fractional differentiation, and then uses the Monte Carlo method to obtain sample nodes that conform to fractional order. This method generates nodes that are denser near the estimated point and less dense at the starting position, similar to known finite difference methods on non-equidistant or nested grids.  Hence, it can be used effectively to solve fPDEs om irregular domains; moreover, it can allow for parallelization. However, since its expression consists of the functional value of the objective function at the generating sample nodes, it cannot be used in the numerical solution of fractional PDEs in classical methods.

Physics-informed neural network (PINNs) \cite{raissi2019}, is an algorithm that combines physical models with deep learning, and has been widely used in solving PDEs \cite{raissi2020, mao2020, jin2021, fpinn, wang2023}. One of its many advantages lies in the ability to construct solutions directly using neural networks (NNs), and then use the loss function to correct the known labeled data in combination with the governing equations to make it an approximate solution that truly satisfies the known conditions. Therefore, with the help of PINNs, we can  apply this new Monte Carlo method effectively to solve fPDEs.
In previous work \cite{lingguo}, the authors combined the Monte Carlo method and fPINN to solve the forward and inverse problems involving fPDEs. However, they only used the Monte Carlo method to approximate the integral in the fractional differentiation for the Caputo fractional derivatives, but this apparoach cannot be directly applied to other cases, e.g., the Riemann-Liouville fractional derivatives. Moreover, it uses unbiased estimators for the loss function, which requires the setting of a range of priori parameter distributions. While this can reduce the computational cost, it also increases the reliance on a priori experience. Therefore, it would also be interesting to develop a more general MC method that does not rely on manual and empirical settings.

In this paper, we propose a new Monte Carlo PINNs method for solving fPDEs on irregular domains. We first extend the Monte Carlo method to the approximation of the fractional derivatives of the right-sided, develop two different quasi-Monte Carlo methods to implement it, and successfully apply it to simulate numerical solutions of fPDEs with the help of PINNs. Compared to the original fPINN method, Monte Carlo PINN uses sample nodes that are block-distributed, and the approximate expression is the average of their corresponding function values, so it is able to achieve a reduction in storage requirements and computational costs by organizing its generation points, which enables it to have better computational efficiency. Moreover, since its node distribution is similar to finite difference methods on non-equidistant or nested grids, this allows our method to better adapt to the shape of the boundary for irregular boundary problems, thus ensuring the accuracy of numerical simulations. We present several examples that further demonstrate the effectiveness of our method. In particular, we test a very interesting numerical example, the fuzzy boundary location problem, that is, we obtain the fPDE solution when the boundary region is uncertain. This problem often arises in situations where it is difficult to accurately measure the precise location of a domain. We compared the numerical results of our method with those of fPINN, and found that our method is able to achieve the same or even better numerical accuracy while guaranteeing higher computational efficiency. Finally, we simulated the real problem of human cerebrovascular blood flow and compared the results with those simulated by traditional methods, which again proved the validity of our method.

The rest of the paper is organized as follows. Section \ref{sec2} presents the general form of fPDEs. Extended Monte Carlo methods and the new Monte Carlo PINNs method are described in detail in Section \ref{sec3}. Section \ref{sec4} presents the numerical examples to verify the effectiveness of our method for fPDEs on irregular domains. Finally, some conclusions are summarized in Section \ref{sec5}.

\section{Problem statement}\label{sec2}
The general form of fPDEs can be expressed as
\begin{equation}
\label{eq.11}
    \begin{aligned}
        &_0^C D_t^{\alpha} u (\mathbf{x}, t) + \mathcal{L}(u, u_{\mathbf{x}\mathbf{x}}, (-\triangle)^{\frac{\beta}{2}}u, ...; \mathbf{x}, t) = f(\mathbf{x}, t),~\mathbf{x}\in\Omega,~t\in(0,T],\\
        &u(\mathbf{x}, 0) = u_0 (\mathbf{x}),~\mathbf{x}\in\Omega,\\
        &u(\mathbf{x}, t) = u (\mathbf{x}, t),~\mathbf{x}\in\partial\Omega,~t\in(0,T],
    \end{aligned}
\end{equation}
where $u (\mathbf{x}, t)$ represents the exact solution, $f(\mathbf{x}, t)$ is the forcing term, and $_0^C D_t^{\alpha}$ represents the Caputo time fractional derivative operator, defined as \cite{george_book}
\begin{equation}
\label{eq.12}
    ^C D_t^{\alpha} u(\mathbf{x}, t) := \frac{1}{\Gamma(1-\alpha)}\int_0^t (t-s)^{-\alpha} u^{(n)}(\mathbf{x}, s)ds,~n-1<\alpha<n,
\end{equation}
and $(-\triangle)^{\frac{\beta}{2}}$ is the Riesz fractional derivative operator \cite{george_book}
\begin{equation}
    (-\triangle)^{\frac{\beta}{2}}u(\mathbf{x}, t) := c_{\beta}(_{\mathbf{x}_{lb}}^{RL}D_{\mathbf{x}}^{\beta}u + \emph{}_{\mathbf{x}}^{RL}D_{\mathbf{x}_{ub}}^{\beta}u),
\end{equation}
with $m-1<\beta<m$, $c_{\beta} = -\frac{1}{2\mathrm{cos}(\pi\beta/2)}$ and
\begin{equation}
    \begin{aligned}
        &_{\mathbf{x}_{lb}}^{RL}D_{\mathbf{x}}^{\beta}u(\mathbf{x}, t) := \frac{1}{\Gamma(m-\beta)}\frac{d^m}{dx^m}\int_{\mathbf{x}_{lb}}^{\mathbf{x}} (\mathbf{x} - \tau)^{m-\beta-1} u(\mathbf{x}+(\tau-\mathbf{x}), t)d\tau,\\
        &_{\mathbf{x}}^{RL}D_{\mathbf{x}_{ub}}^{\beta}u(\mathbf{x}, t) := \frac{(-1)^m}{\Gamma(m-\beta)}\frac{d^m}{dx^m}\int_{\mathbf{x}}^{\mathbf{x}_{ub}} (\tau - \mathbf{x})^{m-\beta-1} u(\mathbf{x}+(\tau-\mathbf{x}), t)d\tau,
    \end{aligned}
\end{equation}
where $\mathbf{x}_{lb}, \mathbf{x}_{ub}$ are the lower and upper bounds of $\Omega$ with respect to $\mathbf{x}$,
\begin{equation}
    \mathbf{x}_{lb} := \mathrm{inf}\{\mathbf{x}:(\mathbf{x},t)\in\Omega\times(0,T]\},~~\mathbf{x}_{ub} := \mathrm{sup}\{x:(\mathbf{x},t)\in\Omega\times(0,T]\}.
\end{equation}
In case of high-dimensional spaces, $\{_{\mathbf{x}_{lb}}^{RL}D_{\mathbf{x}}^{\beta}u\}$ and $\{_{\mathbf{x}}^{RL}D_{\mathbf{x}_{ub}}^{\beta}u\}$ mean $\{_{x_{lb}}^{RL}D_{x}^{\beta}u$, $_{y_{lb}}^{RL}D_{y}^{\beta}$
$u$, $\emph{}_{z_{lb}}^{RL}D_{z}^{\beta}u\}$ and $\{\emph{}_{x}^{RL}D_{x_{ub}}^{\beta}u$, $\emph{}_{y}^{RL}D_{y_{ub}}^{\beta}u$, $\emph{}_{z}^{RL}D_{z_{ub}}^{\beta}u\}$, respectively. $\mathcal{L}(\cdot)$ is a linear or nonlinear space differential operator containing possible orders of space derivatives. For simplicity, here we take $\mathcal{L}(\cdot) := (-\triangle)^{\beta/2}$ as an example. By applying the new Monte Carlo PINNs, we will solve fPDEs, in the form above but not limited to this, on irregular domains.

\section{The General Monte Carlo PINNs Method}\label{sec3}
In this section, we introduce a new Monte Carlo PINNs to apply it to Eq. \eqref{eq.11}. Before describing the details of the method, we first extend the Monte Carlo method to approximate right-sided fractional differentiation.

\subsection{Monte Carlo method for right-sided fractional  differentiation}\label{sec3.1}

The Monte Carlo method for fractional differentiation \cite{podlubny2022,podlubny20222} is a new approximation method. This method defines a sequence of fractional order dependent probability distribution functions based on the Gr\"{u}nwald-Letnikov defined fractional differential operators. It then uses the Monte Carlo method for nodal sampling based on this probability distribution function to construct approximate expressions for fractional differentiation at different fractional orders. Under this idea, an approximate expression for the left-sided fractional-order differentiation of $f(x)$ can be written as
\begin{equation}\label{mc_left}
    \begin{aligned}
         &_{x_{lb}}D_x^{\alpha}f(x) \sim A_{h}^{\alpha} f(x) \\
         &= \frac{1}{K}(\frac{1}{h^{\alpha - n}}(\frac{1}{h^n}\sum_{j=0}^n (-1)^j C_n^j f(x-jh) - \frac{1}{N}\sum_{m=1}^N \frac{1}{h^n} \sum_{j=0}^n (-1)^j C_n^j f(x-(Y_m^{(\alpha)} + j)h))),
    \end{aligned}
\end{equation}
where $x_{lb}$ represents the left bound of the domain with respect to $x$; $\alpha$ is the order of the fractional derivative, $\alpha\in(n,n+1)$; $h$ is the step-size, $h=(x-x_{lb})/N$, where $N$ is the number of approximate points; $K$ is the repeat time and $\{Y_m^{(\alpha)},m=1,...,N\}$ are the sample nodes on $\alpha$ generated by Monte Carlo method; and   $C_n^j$ represents binomial coefficient. This method can be used for approximation of all three types of fractional derivatives: the Caputo, the Gr\"{u}nwald-Letnikov, and the Riemann-Liouville fractional derivatives.

In the original paper \cite{podlubny20222} only the expression for approximating the left-sided fractional derivative is presented.  Therefore, we should derive an approximation of the right-sided fractional derivatives in order to be able to approximate the Riesz fractional derivative operator. As an example, we only show the derivation process of $\alpha\in(1,2)$. The derivation steps are also applicable to other values of $\alpha$.

Similarly, let us first recall that the Gr\"{u}nwald-Letnikov fractional derivative on $L_1 (\mathbb{R})$ defined as \cite{george_book},
\begin{equation}
    _x^{GL}D_{x_{ub}}^{\alpha}f(x) = \underset{h\longrightarrow 0}{\mathrm{lim}}\mathcal{A}_{-h}^{\alpha} f(x),~\alpha>0,~h>0,
\end{equation}
where
\begin{equation}
    \begin{aligned}
        &\mathcal{A}_{-h}^{\alpha}f(x) = \frac{1}{h^{\alpha}}\sum_{k=0}^{\infty} (-1)^{k} \frac{\alpha(\alpha-1)\cdot\cdot\cdot(\alpha-k+1)}{k!}f(x+kh),
    \end{aligned}
\end{equation}
where $x_{ub}$ represents the right boundary of the domain with respect to $x$, $h$ is the step size, $h = (x_{ub} - x)/N$, and  $N$ is the number of approximation points.

We set
\begin{equation}
    m_k = \nu(\alpha,k) = (-1)^k \frac{\Gamma(\alpha+1)}{k!\Gamma(\alpha-k+1)},
\end{equation}
thus
\begin{equation}
    m_0 = 1,~m_1 = \nu(\alpha,1) = -\alpha<0,~\text{and}~m_k > 0,~\text{for}~k=2,3,....
\end{equation}
In addition, the coefficient $m_k$ can be regarded as the series expansion coefficient of $(1-z)^{\alpha}$, $\sum_{k=0}^{\infty} m_k = 0$. Thus, $\sum_{k=1}^{\infty} (-\nu(\alpha,k))=1$.

Since
\begin{equation}
    \begin{aligned}
        \sum_{k=0}^{\infty} m_k &= 0 = 1 - \alpha + \sum_{k=2}^{\infty}m_k,\\
        &1 = 2- \alpha + \sum_{k=2}^{\infty} m_k,\\
    \end{aligned}
\end{equation}
let $Y\in{1,2,...}$ be a discrete random variable such that
\begin{equation}\label{gailv}
    \begin{aligned}
        &P(Y=1) = p_1 = p_1(\alpha) = 2-\alpha\in(0,1),\\
        &P(Y=k) = p_k = p_k(\alpha) = m_k = (-1)^k \frac{\Gamma(\alpha+1)}{k!\Gamma(\alpha-k+1)}>0,~k=2,3,...,
    \end{aligned}
\end{equation}

thus,
\begin{equation}
    \sum_{k=1}^{\infty} p_k = 1.
\end{equation}
Note that $E(Y) < \infty$, but $Var(Y) = \infty$.

Then, we have
\begin{equation}
    \begin{aligned}
        \sum_{k=0}^{\infty} m_k f(x+kh) &= f(x) + \sum_{k=1}^{\infty}m_k f(x+kh)\\
        &= f(x) - \alpha f(x+h) + \sum_{k=2}^{\infty}m_k f(x+kh)\\
        &= f(x) + (-2+p_1(\alpha)) f(x+h) + \sum_{k=2}^{\infty}m_k f(x+kh)\\
        &= f(x) - 2f(x+h) + \sum_{k=1}^{\infty}m_k f(x+kh)\\
        &= f(x) - 2f(x+h) + E f(x+kh),
    \end{aligned}
\end{equation}
if the stochastic process $\xi_h (x) = f(x+Yh)$ is such that for a fixed $f,t$, and $h$, $Ef(x+Yh)<\infty$.

Let $Y_1,Y_2,...,Y_n,...$ be independent copies of the random variable $Y$, then by the strong law of large numbers
\begin{equation}
    \frac{1}{N}\sum_{n=1}^N f(x+Y_n h)\longrightarrow Ef(x+Yh),~N\longrightarrow\infty,
\end{equation}
with probability one for any fixed $t$ and $h$. Thus,
\begin{equation}\label{mc_right}
    \mathcal{A}_{N,-h}^{\alpha} = \frac{1}{h^{\alpha}}[f(x) - 2f(x+h) + \frac{1}{N}\sum_{n=1}^N f(x+Y_n h)],
\end{equation}
with probability one converges to $\mathcal{A}_{-h}^{\alpha}f(x),\alpha\in(1,2)$. Furthermore, same as \cite{podlubny20222}, we have
\begin{equation}
    \mathcal{A}_{N,-h}^{\alpha} = \frac{1}{K}(\frac{1}{h^{\alpha}}(f(x) - 2f(x+h) + \frac{1}{N}\sum_{n=1}^N f(x+Y_n h))).
\end{equation}

\begin{figure}[htb]
\centering
\subfigure[] 
{
	\begin{minipage}[t]{0.3\linewidth}
	\includegraphics[width=1\textwidth]{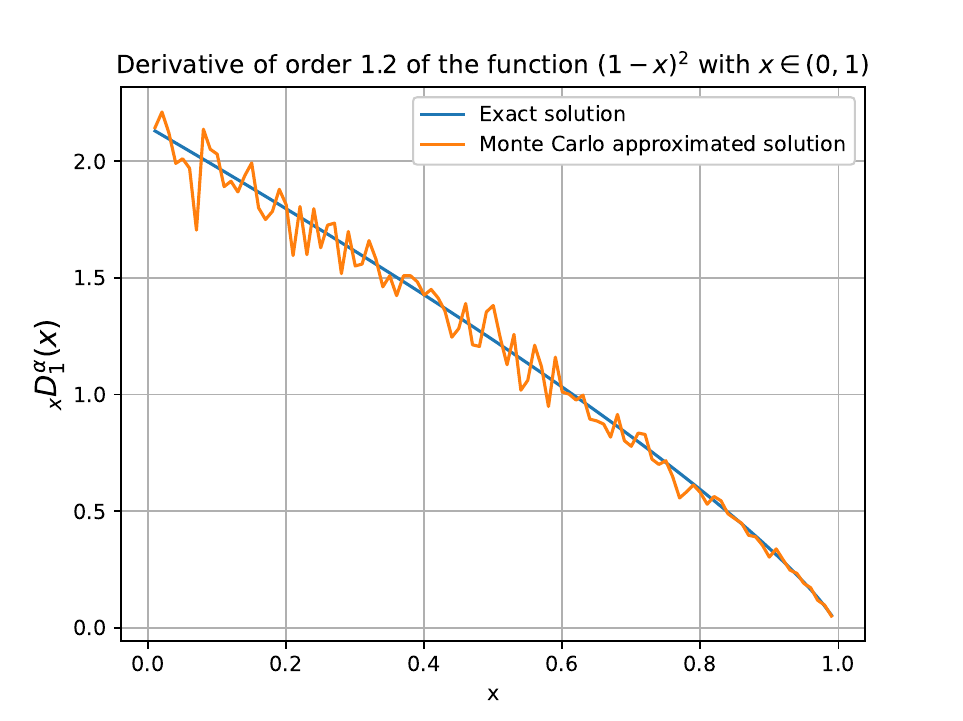}   
	\end{minipage}
}
\subfigure[] 
{
	\begin{minipage}[t]{0.3\linewidth}
	\includegraphics[width=1\textwidth]{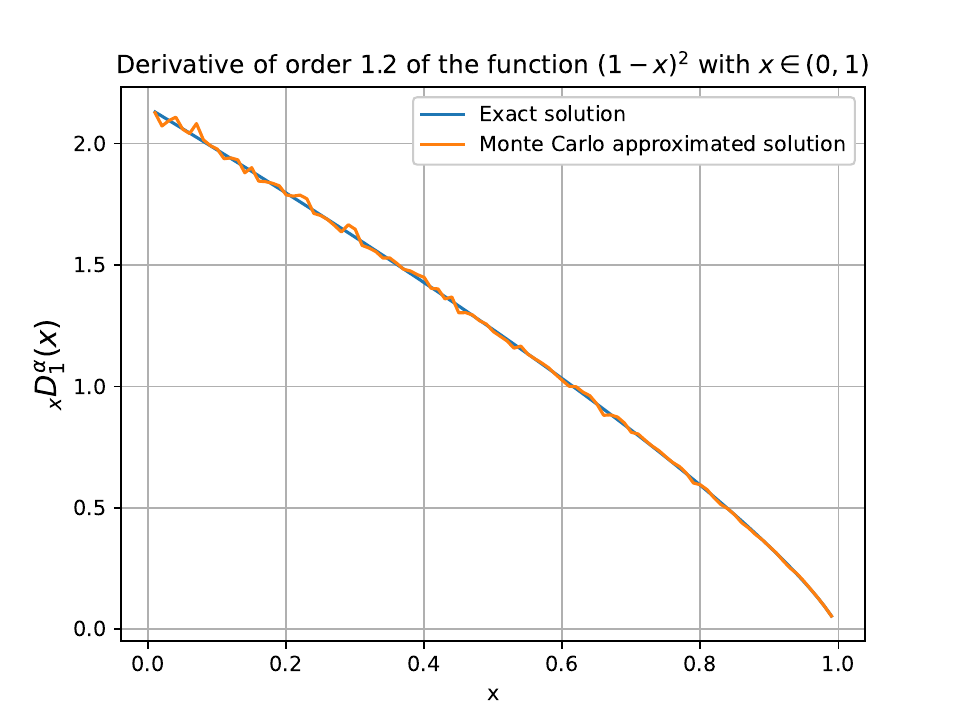}   
	\end{minipage}
}
\subfigure[] 
{
	\begin{minipage}[t]{0.3\linewidth}
	\includegraphics[width=1\textwidth]{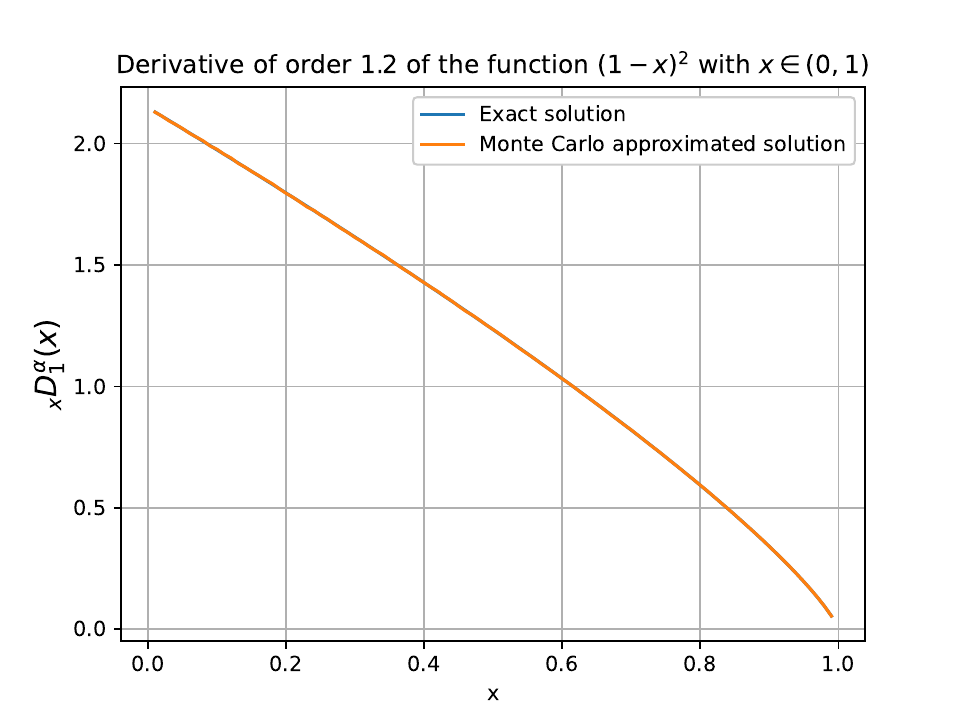}   
	\end{minipage}
}
\centering	
\subfigure[] 
{
	\begin{minipage}[t]{0.3\linewidth}
	\includegraphics[width=1\textwidth]{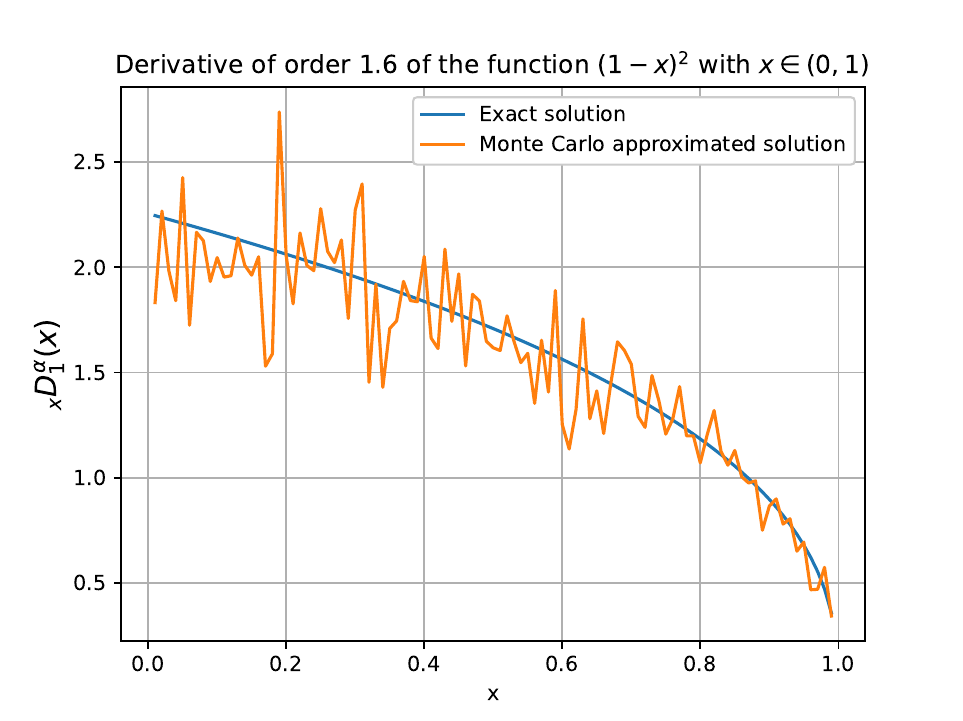}   
	\end{minipage}
}
\subfigure[] 
{
	\begin{minipage}[t]{0.3\linewidth}
	\includegraphics[width=1\textwidth]{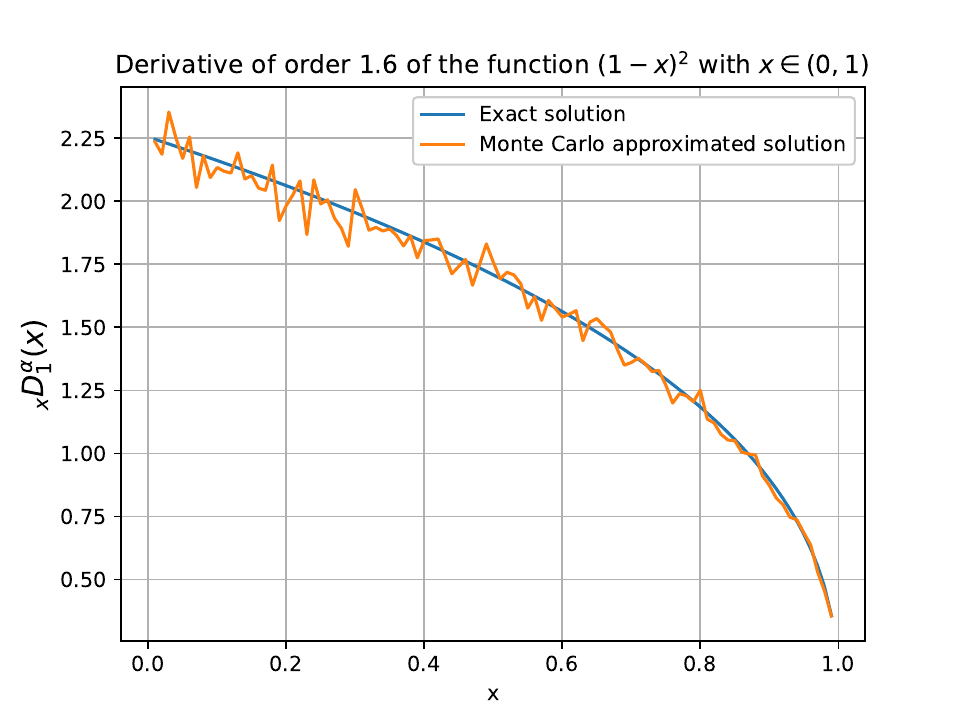}   
	\end{minipage}
}
\subfigure[] 
{
	\begin{minipage}[t]{0.3\linewidth}
	\includegraphics[width=1\textwidth]{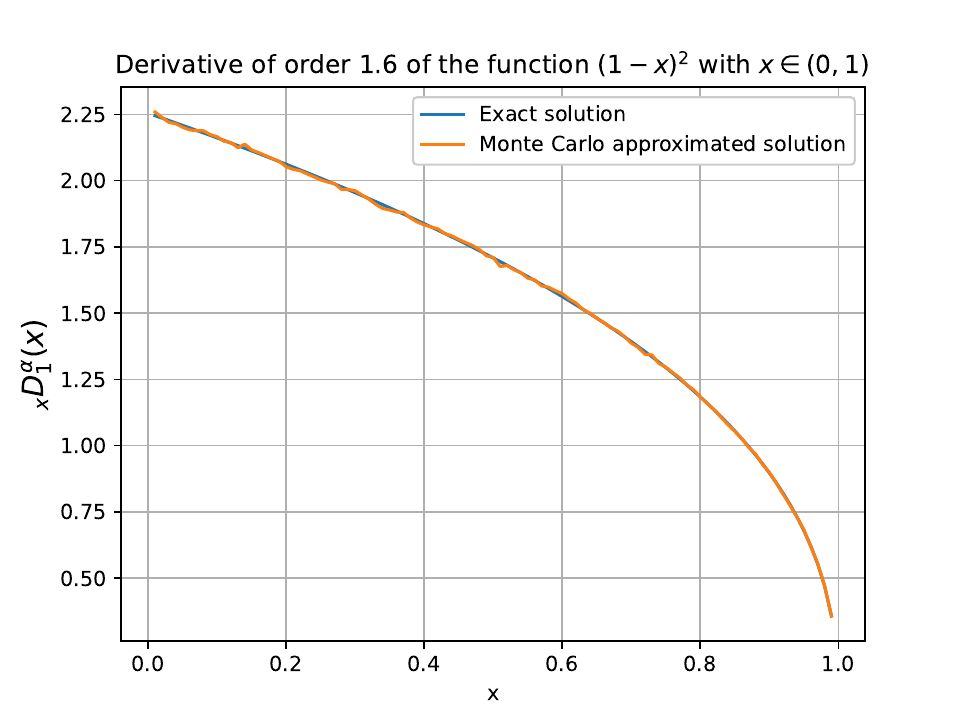}   
	\end{minipage}
}
\caption{Results of the Monte Carlo method of the right-sided fractional derivatives of the function $(1-x)^2$ at different numbers of $N, K$ for different fractional orders $\alpha$. For the case $\alpha = 1.2$: (a) $K = N = 10$, $L^2~\mathrm{error} = 5.63\mathrm{e}^{-2}$. (b) $K = N = 100$, $L^2~\mathrm{error} = 9.94\mathrm{e}^{-3}$. (c) $K = N = 3000$, $L^2~\mathrm{error} = 6.1\mathrm{e}^{-5}$. For the case $\alpha = 1.6$: (d) $K = N = 10$, $L^2~\mathrm{error} = 1.11\mathrm{e}^{-1}$. (e) $K = N = 100$, $L^2~\mathrm{error} = 2.61\mathrm{e}^{-2}$. (f) $K = N = 3000$, $L^2~\mathrm{error} = 9.95\mathrm{e}^{-5}$.} 
\label{fig2}  
\end{figure}

Next, we replace the sample $Y_i$ by its Monte Carlo simulations. Define
\begin{equation}
    E_j = \sum_{i=1}^j p_i,
\end{equation}
where $p_i = p_i(\alpha)$, then
\begin{equation}
0=E_0<E_1<E_2<...<E_j<...,~with~p_i=E_i-E_{i-1}.
\end{equation}
Here, we take $F$ as a random variable uniformly distributed on $[0,1]$, then
\begin{equation}
    P(E_{j-1}<F<E_j) = p_j,
\end{equation}
and to generate $Y\in1,2,...$ we set $Y=k$ if $E_{k-1}\leq F<E_k$. The right-sided fractional derivatives of the function $(1-x)^2$ for different fractional orders $\alpha$ are presented  in Fig. \ref{fig2} to demonstrate the correctness of our method, where $x\in(0,1)$.

As can be seen from Fig. \ref{fig2}, the fit gets better as $N$ and $K$ increase, proving the validity of our method to approximate the right-sided fractional derivatives. In addition, it is worth noting that $F$ can also be selected as a low discrepancy sequence. Next, we test the Sobol and Halton sequences separately and give the corresponding steps of the algorithm.
\begin{figure}[htbp]
\centering
\subfigure[] 
{
	\begin{minipage}[t]{0.22\linewidth}
	\includegraphics[width=1\textwidth]{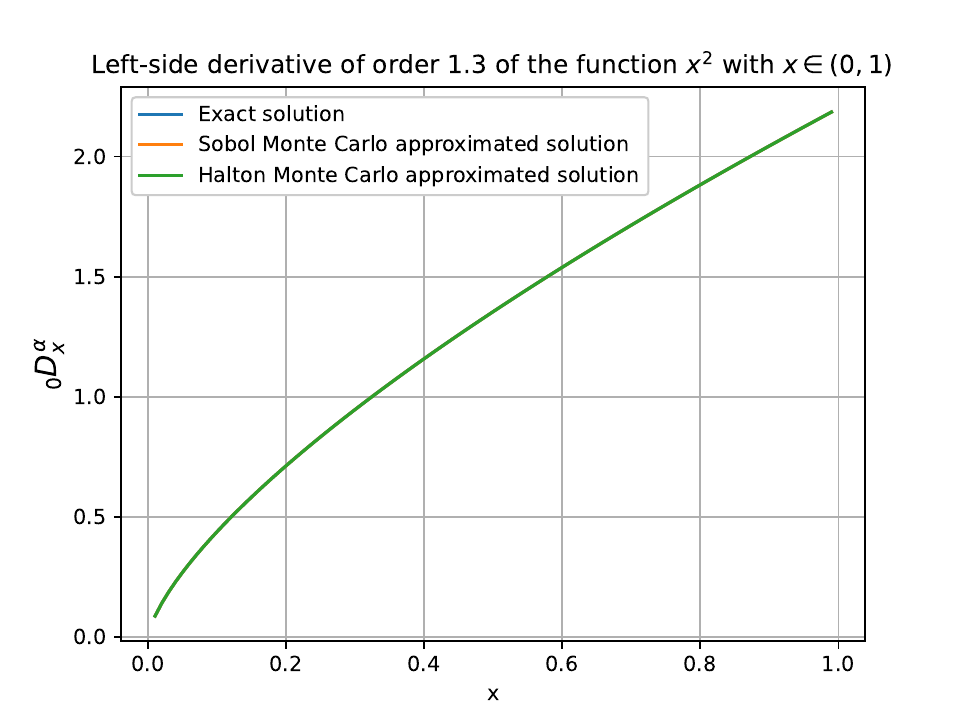}   
	\end{minipage}
}
\subfigure[] 
{
	\begin{minipage}[t]{0.22\linewidth}
	\includegraphics[width=1\textwidth]{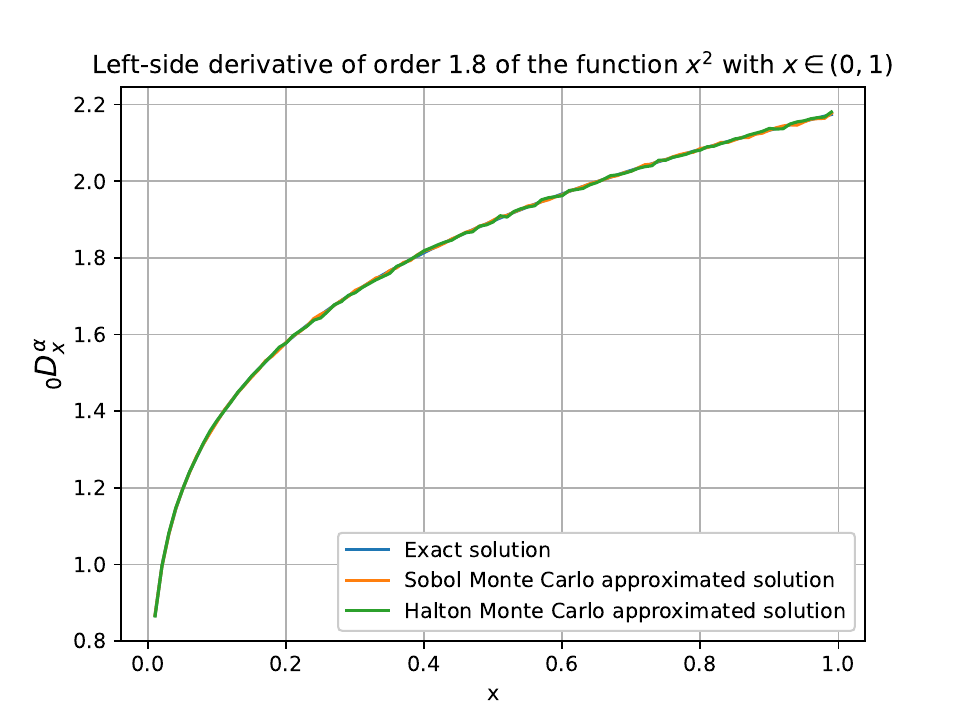}   
	\end{minipage}
}
\subfigure[] 
{
	\begin{minipage}[t]{0.22\linewidth}
	\includegraphics[width=1\textwidth]{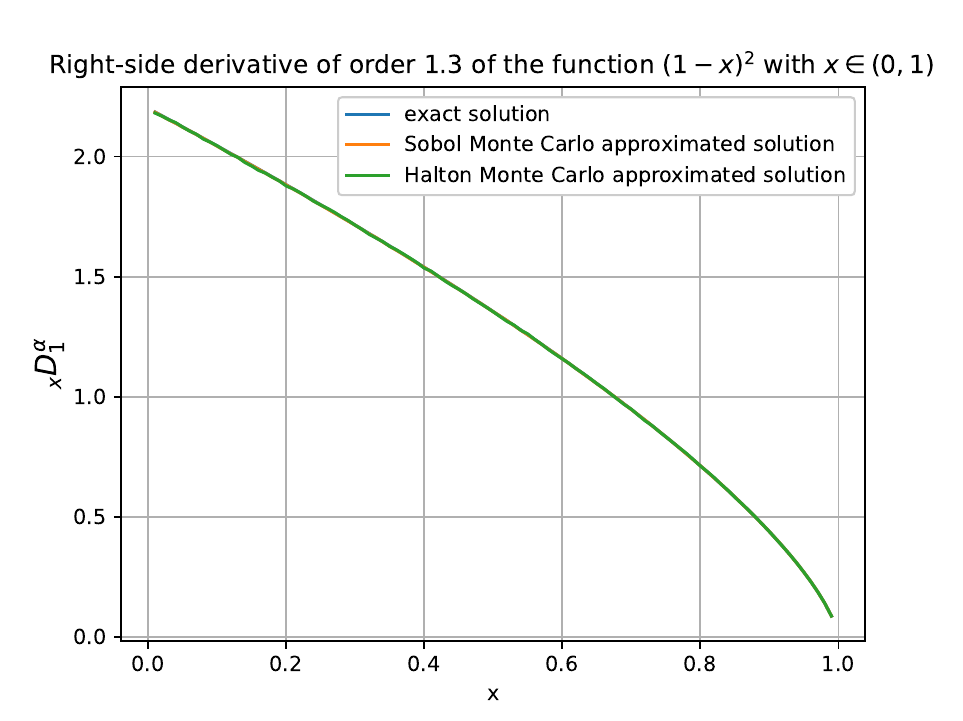}   
	\end{minipage}
}
\subfigure[] 
{
	\begin{minipage}[t]{0.22\linewidth}
	\includegraphics[width=1\textwidth]{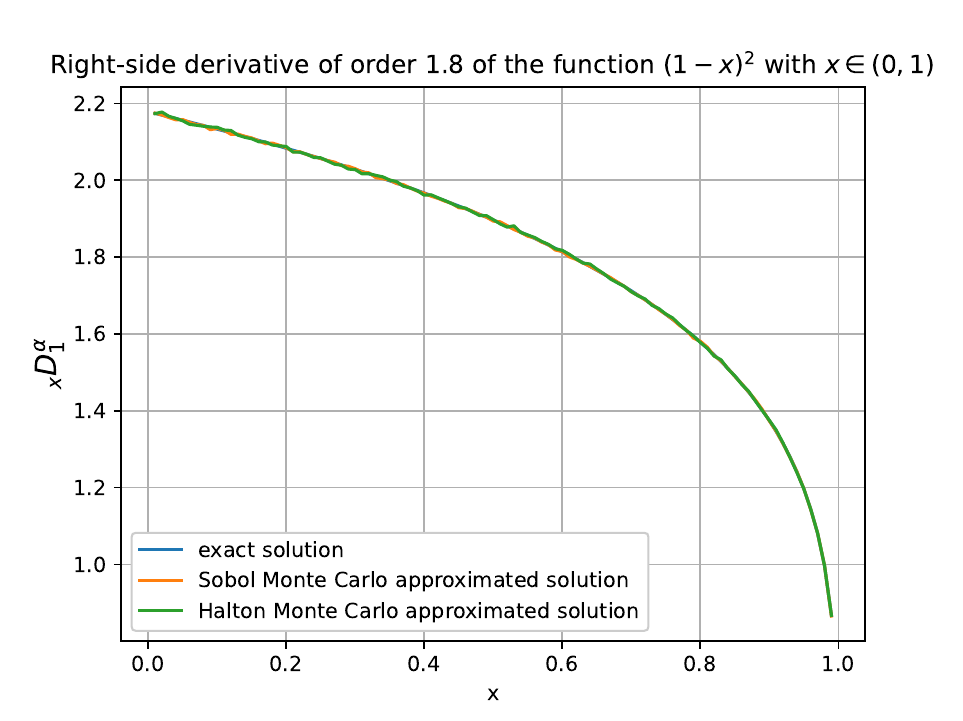}   
	\end{minipage}
}
\caption{Results of the Quasi-Monte Carlo method (Sobol/Halton) of the left-sided and right-sided fractional derivatives of $x^2$ and $(1-x)^2$ at $N = K = 1000$ for different fractional orders $\alpha$. For left-sided fractional derivative of $x^2$: (a) $\alpha = 1.3$, $L^2~\mathrm{error} = 4.51\mathrm{e}^{-4}$ and $L^2~\mathrm{error} = 4.47\mathrm{e}^{-4}$. (b) $\alpha = 1.8$, $L^2~\mathrm{error} = 5.07\mathrm{e}^{-4}$ and $L^2~\mathrm{error} = 4.91\mathrm{e}^{-4}$. For right-sided fractional derivative of $(1-x)^2$: (c) $\alpha = 1.3$, $L^2~\mathrm{error} = 4.49\mathrm{e}^{-4}$ and $L^2~\mathrm{error} = 4.18\mathrm{e}^{-4}$. (d)$ \alpha = 1.8$, $L^2~\mathrm{error} = 4.96\mathrm{e}^{-4}$ and $L^2~\mathrm{error} = 4.88\mathrm{e}^{-4}$.} 
\label{fig3}  
\end{figure}

\subsection{Quasi-Monte Carlo method for fractional  differentiation}\label{sec3.2}
In this subsection, we show how to approximate the fractional differentiation using the quasi-Monte Carlo method. The difference with the Monte Carlo method is that the samples are selected in low-discrepancy sequences instead of random sampling. The two main steps of the algorithm are as follows.

\textbf{Step 1}. Defining $\widetilde{E}_j = \sum_{i=1}^j p_i$, where $p_i = p_i(\alpha)$ (same as Eq. \eqref{gailv}), then $0=\widetilde{E}_0<\widetilde{E}_1<\widetilde{E}_2<...<\widetilde{E}_j<...,~with~p_i=\widetilde{E}_i-\widetilde{E}_{i-1}$;

\textbf{Step 2}.
Taking $\hat{F}$ is a chosen random variable on the Sobol sequence (or the Halton sequence), then
\begin{equation}
    P(\widetilde{E}_{j-1}<\hat{F}<\widetilde{E}_j) = p_j,
\end{equation}
and to generate $Y\in1,2,...$ we set $Y=k$ if $\widetilde{E}_{k-1}\leq \hat{F}<\widetilde{E}_k$.

We have similarly tested the above methods numerically. In Fig. \ref{fig3}, the Sobol and Halton Monte Carlo methods are used to approximate the left-sided and right-sided fractional derivatives of $x^2$ and $(1-x)^2$ at different fractional orders, respectively. The numerical results prove the effectiveness of the above method, and by comparing with the exact numerical results, it can be found that the quasi-Monte Carlo method can achieve better convergence accuracy than Monte Carlo method at smaller $N$ and $K$.
\begin{figure}[htbp]
\centering
\includegraphics[width=1\linewidth]{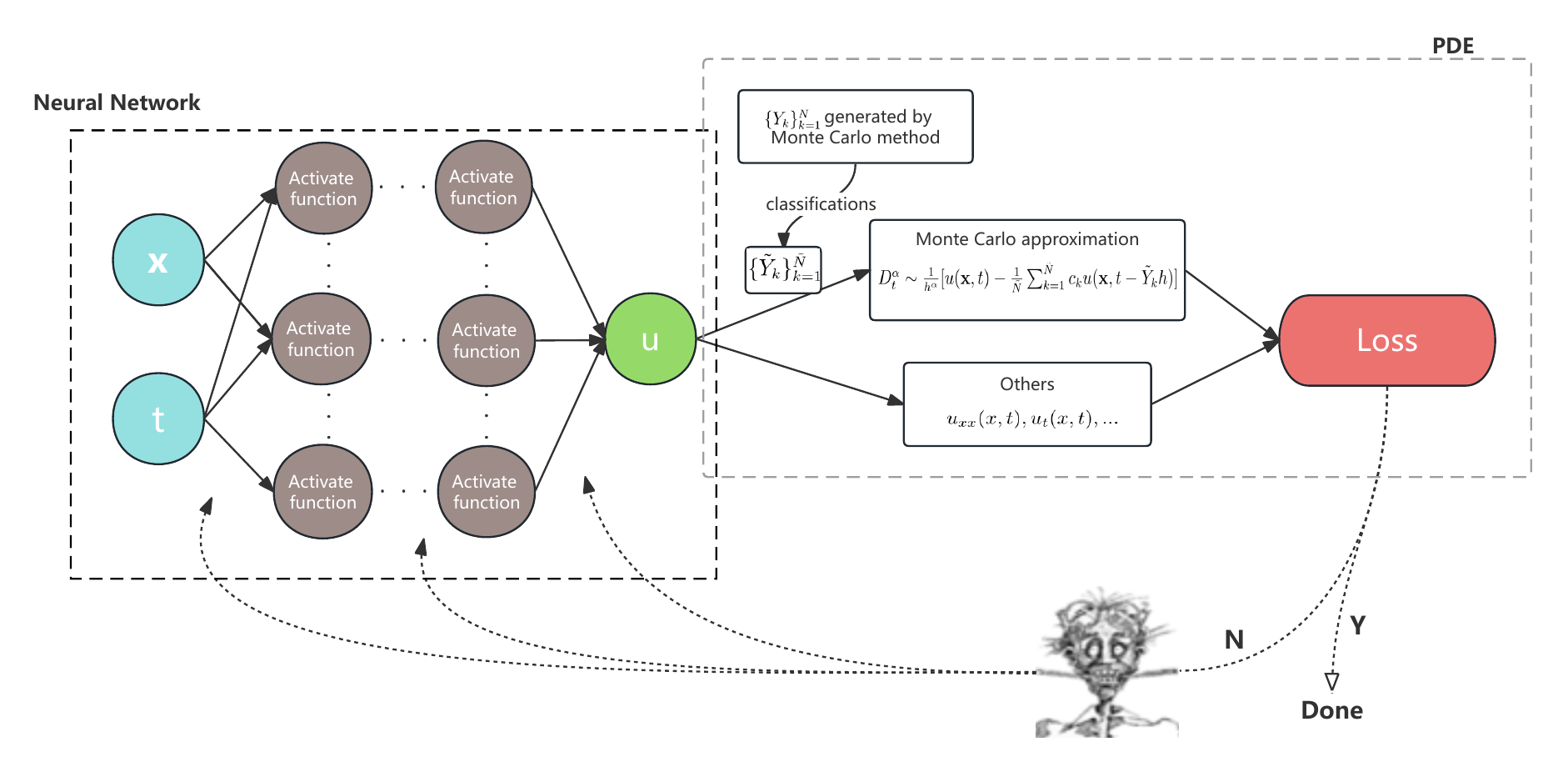}\\
\centering
\caption{Schematic of the General Monte Carlo PINN method.}
\label{fig1}
\end{figure}

\subsection{The PINN Solver}\label{sec3.3}
Based on the approximation formula for fractional-order differentiation given in the previous subsection, a new Monte Carlo PINN is introduced for solving fPDEs on irregular domains in this subsection, as shown in Fig. \ref{fig1}. Here, we take Eq. \eqref{eq.11} as an example and set $\alpha\in(0,1)$ and $\beta\in(1,2)$.

Using Monte Carlo approximations \eqref{mc_left} and \eqref{mc_right} for the left- and right-sided derivatives, we can get the following fully discrete scheme for the left-hand-side of Eq. \eqref{eq.11}.
\begin{equation}
    \begin{aligned}
        L^* &= \frac{1}{K_t}[\frac{1}{h_t^{\alpha}}[u(\mathbf{x}, t) - \frac{1}{N_t}\sum_{m=1}^{N_t} f(\mathbf{x}, t-Y_m^{(t)} h_t)]] \\
        &+ c_\beta\{\frac{1}{^{lb}K_{\mathbf{x}}}(\frac{1}{{}^{lb}h_{\mathbf{x}}^{\beta}}[u(\mathbf{x}, t) - 2u(\mathbf{x}-{}^{lb}h_{\mathbf{x}}, t) + \frac{1}{^{lb}N_{\mathbf{x}}}\sum_{n=1}^{^{lb}N_{\mathbf{x}}} u(\mathbf{x}-Y_n^{(\mathbf{x}_{lb})} {}^{lb}h_{\mathbf{x}}, t)]) \\
        &+ \frac{1}{^{ub}K_{\mathbf{x}}}(\frac{1}{^{ub}h_{\mathbf{x}}^{\beta}}[u(\mathbf{x}, t) - 2u(\mathbf{x}+^{ub}h_{\mathbf{x}}, t) + \frac{1}{^{ub}N_{\mathbf{x}}}\sum_{j=1}^{^{ub}N_{\mathbf{x}}} u(\mathbf{x}+Y_j^{(\mathbf{x}_{ub})} {}^{ub}h_{\mathbf{x}}, t)])\},
    \end{aligned}
\end{equation}
where $K_t$, $^{lb}K_{\mathbf{x}}$, $^{ub}K_{\mathbf{x}}$ represent the repeat times;  $h_t$, $^{lb}h_{\mathbf{x}}$, $^{ub}h_{\mathbf{x}}$ are step sizes with respect to $t$, $\mathbf{x}$; $h_t = t/N_t$, $^{lb}h_{\mathbf{x}} = (\mathbf{x} - \mathbf{x}_{lb})/^{lb}N_{\mathbf{x}}$, and $^{ub}h_{\mathbf{x}} = (\mathbf{x}_{ub} - \mathbf{x})/^{ub}N_{\mathbf{x}}$; $\{Y_m^{(t)}\}_{m=1}^{N_t}, \{Y_n^{(\mathbf{x}_{lb})}\}_{n=1}^{^{lb}N_\mathbf{x}},$ and$\{Y_j^{(\mathbf{x}_{ub})}\}_{j=1}^{^{ub}N_\mathbf{x}}$ represent the samples generated by Monte Carlo method for $_0D_t^{\alpha},{}_{\mathbf{x}_{lb}}D_\mathbf{x}^{\beta},$ and$ _{\mathbf{x}}D_{\mathbf{x}_{ub}}^{\beta}$, respectively.

\begin{figure}[htbp]
\centering
\subfigure[] 
{
	\begin{minipage}[t]{0.9\linewidth}
	\includegraphics[width=1\textwidth]{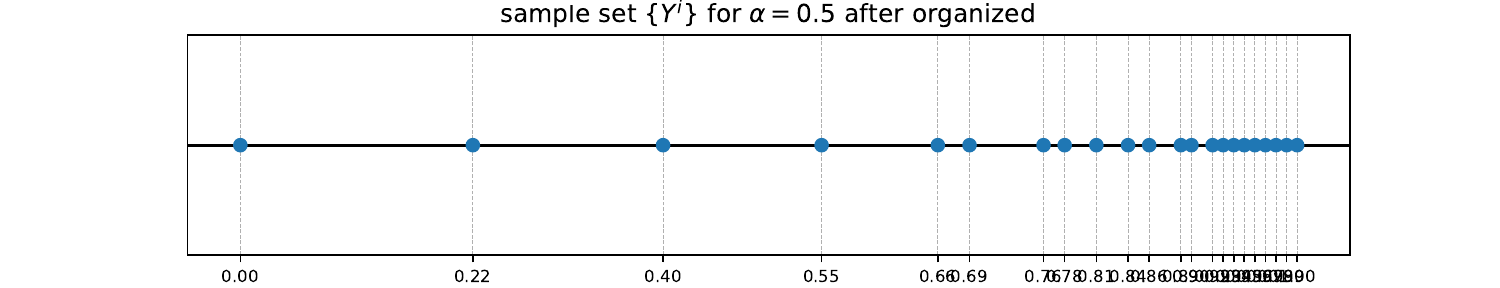}   
	\end{minipage}
}
\subfigure[] 
{
	\begin{minipage}[t]{0.9\linewidth}
	\includegraphics[width=1\textwidth]{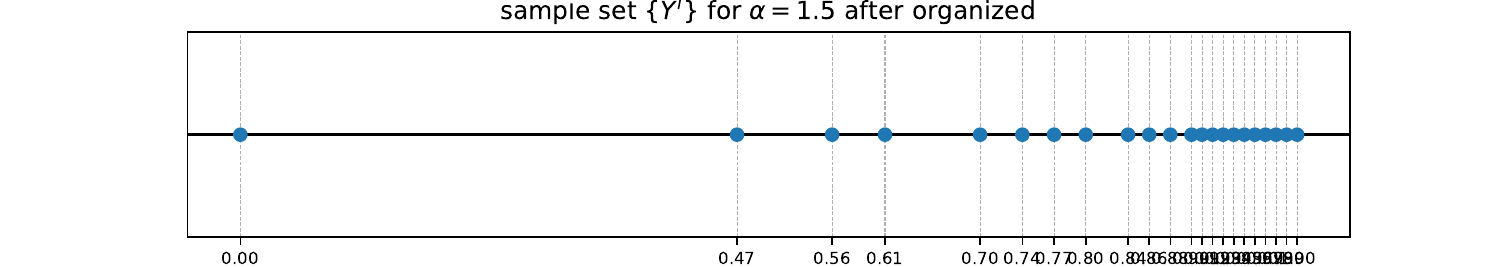}   
	\end{minipage}
}
\caption{The generating point distribution of the reduced set $\{Y_i\}$. Using the Monte Carlo method to generate a total of 100 approximation points reduces to 21, 22 respectively, for (a) $\alpha=0.5$; (b) $\alpha=1.5$, after classification.} 
\label{figy}  
\end{figure}

It is worth noting that since the data points in the sample set $\{Y_i\}$ generated by the Monte Carlo method show a block distribution, we need to perform a simple classification process to reduce the computational cost. The improvement in computational speed of this technique is not obvious when a smaller number of approximation points is chosen, but it becomes more obvious as the number of approximation points increases.  As can be seen in Fig. \ref{figy}, for an approximation with 100 total points, the number can be reduced to about 20 representative  points after classification to eliminate repeats, which means that the storage is reduced by $80\%$ but there are also
substantial computational savings.

Then, we consider the training data set $\mathcal{D} = (\mathcal{D}_{equ}, \mathcal{D}_{ini}, \mathcal{D}_{bound})$. By using the network output as an approximated solution, we can define the loss function as follows,
\begin{equation}
    Loss(\theta=\{weights,biases\}) = \mathcal{\omega}_{E}\mathrm{MSE}_{E} + \mathcal{\omega}_{I}\mathrm{MSE}_{I} + \mathcal{\omega}_{B}\mathrm{MSE}_{B}
\end{equation}
where
\begin{equation}
    \begin{aligned}
        &\mathrm{MSE}_{E} = \frac{1}{N_E}\sum_{i=1}^{N_E}|\mathrm{L^*(u_{NN}}; \mathbf{x}_i, t_i) - f(\mathbf{x}_i, t_i)|^2, \\
        &\mathrm{MSE}_{B} = \frac{1}{N_B}\sum_{i=1}^{N_B}|u_{NN}(\mathbf{x}_i, t_i) - u(\mathbf{x}_i, t_i)|^2,~\mathrm{MSE}_{I} = \frac{1}{N_I}\sum_{i=1}^{N_I}|u_{NN}(\mathbf{x}_i, 0) - u_0(\mathbf{x}_i)|^2.
    \end{aligned}
\end{equation}
Here $\mathcal{\omega}_{E},\mathcal{\omega}_{I}$ and $\mathcal{\omega}_{B}$ are penalty weights, which we choose manually here but a better method is available in \cite{RBA-Juan}. $N_E, N_I$ and $N_B$ represent the number of collocation points, initial condition points and boundary condition points, respectively. An optimization method, e.g. Adam, is then used to find the optimal parameter $\theta$ to minimize the value of the loss function, thus obtaining an approximated solution that satisfies the objective equation.

\section{Numerical experiments}\label{sec4}
In this section, several numerical examples are presented to demonstrate the effectiveness of our method for solving fPDEs on irregular domains. Specifically, the effectiveness of our method in solving the forward problems of the 2D fPDEs on irregular domains is demonstrated by solving the 2D space fractional Poisson equation and the 2D time-space fractional diffusion equation on unit disk and heart-shaped domains, and by comparing the numerical results with the original fPINN method. Numerical results and computational speed are also presented for different number of approximation points. In addition, the forward problem for the 3D coupled time-space fractional Bloch-Torrey equation on ventricular domain of the human brain is solved and the results are compared with classical numerical methods. We test our method on an interesting and practical problem, and the results show the potential of our approach in dealing with fPDEs with fuzzy boundary locations. Finally, numerical results verify the effectiveness and accuracy of our method as well as an application of MRI in the brain.

We express the numerical accuracy in terms of the $L^2$ relative error and point-by-point error, respectively.
\begin{equation}
    L^2~\text{relative error} = \sqrt{\frac{{\sum_i[u(\mathbf{x}_i,t_i) - \hat{u}(\mathbf{x}_i,t_i)]}^{2}}{{\sum_i[u(\mathbf{x}_i,t_i)]}^2}},~\text{Point-wise error} = \frac{{u(\mathbf{x}_i,t_i) - \hat{u}(\mathbf{x}_i,t_i)}}{{u(\mathbf{x}_i,t_i)}}
\end{equation}
where $u$ represents the exact solution and $\hat{u}$ is the approximated solution. Similar to PINNs, the Xavier initialization method \cite{xavier} and the Adam+L-BFGS method \cite{lbfgs} are used to optimize the loss function. Unless stated otherwise, the neural network structure and method parameters used are presented in the Table 1.

\begin{table}[htb]
\renewcommand{\arraystretch}{1}
\label{tab:structure}
\centering
\setlength{\tabcolsep}{0.8mm}{
\begin{tabular}{ccccccccccc}
\hline
\multicolumn{6}{c}{Approximate points $N$}                       & 50    & & &Repeat~times~$K$                & 50    \\ \hline
\multicolumn{6}{c}{Hidden layer}                & 5   &Neurons                & 20    &Activation function                & $\tanh$    \\ \hline
\multicolumn{6}{c}{Optimizer}                & Adam + L-BFGS         &LR                & $1\mathrm{e}^{-3}$    &Iteration                & 30w    \\ \hline
\end{tabular}}
\caption{Parameters used in the numerical results.}
\end{table}

Now we first present Example 4.1 and Example 4.2 to show the effectiveness and accuracy of our method on 2D fPDEs on different geometric domains. We perform numerical experiments for each example separately on the following two different domains:
\begin{equation}
    \left\{
    \begin{aligned}
        &\text{unit disk}~\Omega_1 = ||x-x_{centre}||_2^2 + ||y-y_{centre}||_2^2 \leq 1,\\
        &\Omega_2 = \{(\hat{x},\hat{y}):||\hat{x}||^2 + ||\hat{y}||^2 \leq x^2 + y^2,~\text{where}\{_{y = 1.3cost - 5cos2t - 2cos3t - cos4t,}^{x = 1.6sin^3 t,}
    \end{aligned}
    \right.
\end{equation}
where $t \in [0, 2\pi]$.

\textbf{Example 4.1.} We consider the following 2D space fractional PDEs with initial and boundary conditions:
\begin{equation}\label{examp4.1}
    (-\triangle)_{xx}^{\alpha/2}u(x,y) + (-\triangle)_{yy}^{\beta/2}u(x,y) = f(x,y),~(x,y)\in\Omega,
\end{equation}
with boundary conditions
\begin{equation}
    u(x,y)=u_b (x,y),~(x,y)\in\partial\Omega.
\end{equation}
The fractional operators $(-\triangle)_{xx}^{\alpha/2}$ and $(-\triangle)_{yy}^{\beta/2}$ are defined as
\begin{equation}
    (-\triangle)_{xx}^{\alpha/2} := \frac{1}{2\cos(\pi\alpha/2)}(_{x_{lb}}D_x^{\alpha} + _xD_{x_{ub}}^{\alpha}),~(-\triangle)_{yy}^{\beta/2} := \frac{1}{2\cos(\pi\beta/2)}(_{y_{lb}}D_y^{\beta} + _yD_{y_{ub}}^{\beta}),
\end{equation}
where $_{x_{lb}}D_x^{\alpha}$, $_xD_{x_{ub}}^{\alpha}$, $_{y_{lb}}D_y^{\beta}$ and $_yD_{y_{ub}}^{\beta}$ are the left-sided and right-sided Riemann-Liouville fractional derivatives defined over the $x_{lb}$, $x_{ub}$, $y_{lb}$ and $y_{ub}$, respectively.

The exact solution is $(x-x_{lb})^3 (x_{ub}-x)^3 (y-y_{lb})^3 (y_{ub}-y)^3$, where $x_{lb}$, $x_{ub}$, $y_{lb}$, $y_{ub}$ are the low and upper bound to $x$ and $y$, respectively. The general form of the forcing term is
\begin{equation}
    f(x,y) = f_1(x,y) + f_2(x,y).
\end{equation}
\begin{equation}
    \begin{aligned}
        f_1(x,y) = &\frac{1}{2\cos(\frac{\pi\alpha}{2})}\{\frac{|x_{lb}+x_{ub}|^3 \Gamma(4)}{\Gamma(4-\alpha)}((x - x_{lb})^{3-\alpha} + (x_{ub} - x)^{3-\alpha}) \\
        &- \frac{3 |x_{lb}+x_{ub}|^2 \Gamma(5)}{\Gamma(5-\alpha)}((x - x_{lb})^{4-\alpha} + (x_{ub} - x)^{4-\alpha}) \\
        &+ \frac{3 |x_{lb}+x_{ub}| \Gamma(6)}{\Gamma(6-\alpha)}((x - x_{lb})^{5-\alpha} + (x_{ub} - x)^{5-\alpha}) \\
        &- \frac{3\Gamma(7)}{\Gamma(7-\alpha)}((x - x_{lb})^{6-\alpha} + (x_{ub} - x)^{6-\alpha})\} (y-y_{lb})^3 (y_{ub}-y)^3,
    \end{aligned}
\end{equation}
\begin{equation}
    \begin{aligned}
        f_2(x,y) = &\frac{1}{2\cos(\frac{\pi\beta}{2})}\{\frac{|y_{lb}+y_{ub}|^3 \Gamma(4)}{\Gamma(4-\beta)}((y - y_{lb})^{3-\beta} + (y_{ub} - y)^{3-\beta}) \\
        &- \frac{3 |y_{lb}+y_{ub}|^2 \Gamma(5)}{\Gamma(5-\beta)}((y - y_{lb})^{4-\beta} + (y_{ub} - y)^{4-\beta}) \\
        &+ \frac{3 |y_{lb}+y_{ub}| \Gamma(6)}{\Gamma(6-\beta)}((y - y_{lb})^{5-\beta} + (y_{ub} - y)^{5-\beta}) \\
        &- \frac{3\Gamma(7)}{\Gamma(7-\beta)}((y - y_{lb})^{6-\beta} + (y_{ub} - y)^{6-\beta})\} (x-x_{lb})^3 (x_{ub}-x)^3.
    \end{aligned}
\end{equation}

We first consider the validity of the above equations over the  region $\Omega_1$. In Table 2, we present the errors and computational times of the fPINN method (\cite{fpinn}), the Monte Carlo fPINN method (\cite{lingguo}), and our new (quasi) Monte Carlo PINN method for Example 1 at different approximation points. Compared with the original fPINN method, it can be seen that our method is able to guarantee the same or even better numerical accuracy while exhibiting faster computation speed. Moreover, it can be clearly seen that the convergence accuracy becomes better as the number of approximation points increases, and the computational speed of our method is better compared to fPINN in the case of a larger number of approximation points. Compared to the method in \cite{lingguo}, since we are using the fractional differentiation in 2D space defined by Riemann-Liouville fractional derivatives for this example, we need to use 'tf.gradients' to compute the integer order derivatives outside the integral after using the approximate integrals. This  will cause the computation time to increase substantially with the increase in the number of sample points. However, for irregular boundaries, it is not possible to use any tricks to make the approximate solution automatically satisfy the boundary conditions. Thus, it is not possible to achieve the computational time of the original paper with respect to this problem. Since it is still based on the fPINN method in general, its accuracy does not differ much from that of fPINN. This also shows that our method may have a generalization advantage over the Monte Carlo fPINN of \cite{lingguo}.

\begin{table}[htp]
  \centering
  \fontsize{4}{4}\selectfont
  \begin{threeparttable}
  \caption{Comparison of performance between fPINN method (\cite{fpinn}), the Monte Carlo fPINN method (\cite{lingguo}), and our new (quasi) Monte Carlo PINN method for Example 4.1 on $\Omega_1$ (disk) with $\alpha=1.7, \beta=1.5$. For each method, we set the number of iterations to 300K.}
  \label{tab:performance_comparison}
    \begin{tabular}{ccccccccccccc}
    \toprule
    \multirow{1}{*}{N=K}&
    \multicolumn{2}{c}{ fPINN \cite{fpinn}}&\multicolumn{2}{c}{ Monte Carlo PINN \cite{lingguo}}&\multicolumn{2}{c}{ Sobol Monte Carlo PINN}&\multicolumn{2}{c}{ Halton Monte Carlo PINN}\cr
    \cmidrule(lr){2-3} \cmidrule(lr){4-5} \cmidrule(lr){6-7} \cmidrule(lr){8-9}
    &accuracy&time to run ten iterations&accuracy&time to run ten iterations&accuracy&time to run ten iterations&accuracy&time to run ten iterations\cr
    \midrule
    8&3.74e-01&0.03&2.91e-01&0.03&2.79e-01&0.02&3.57e-01&0.02\cr
    16&1.73e-01&0.05&1.45e-01&0.06&1.35e-01&0.03&1.58e-01&0.03\cr
    32&6.71e-02&0.09&5.97e-02&0.10&5.33e-02&0.05&6.2e-02&0.05\cr
    64&3.97e-02&0.15&1.42e-02&0.17&2.43e-02&0.1&2.57e-02&0.1\cr
    128&9.54e-03&0.26&8.06e-03&0.28&7.79e-03&0.16&8.11e-03&0.16\cr
    \bottomrule
    \end{tabular}
    \end{threeparttable}
\end{table}
\begin{figure}[htb]
\centering
\subfigure[] 
{
	\begin{minipage}[t]{0.4\linewidth}
	\includegraphics[width=1\textwidth]{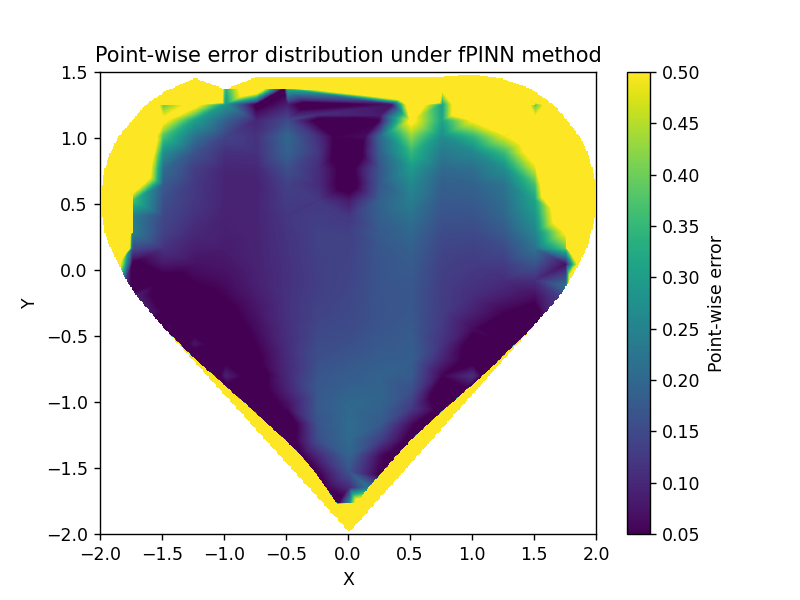}   
	\end{minipage}
}
\subfigure[] 
{
	\begin{minipage}[t]{0.4\linewidth}
	\includegraphics[width=1\textwidth]{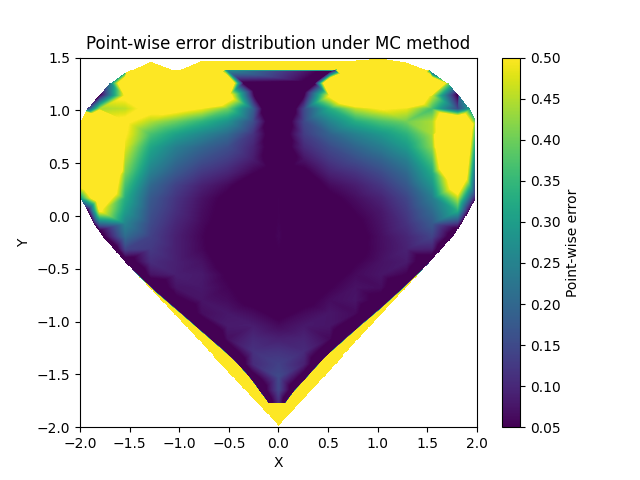}   
	\end{minipage}
}
\centering
\subfigure[] 
{
	\begin{minipage}[t]{0.4\linewidth}
	\includegraphics[width=1\textwidth]{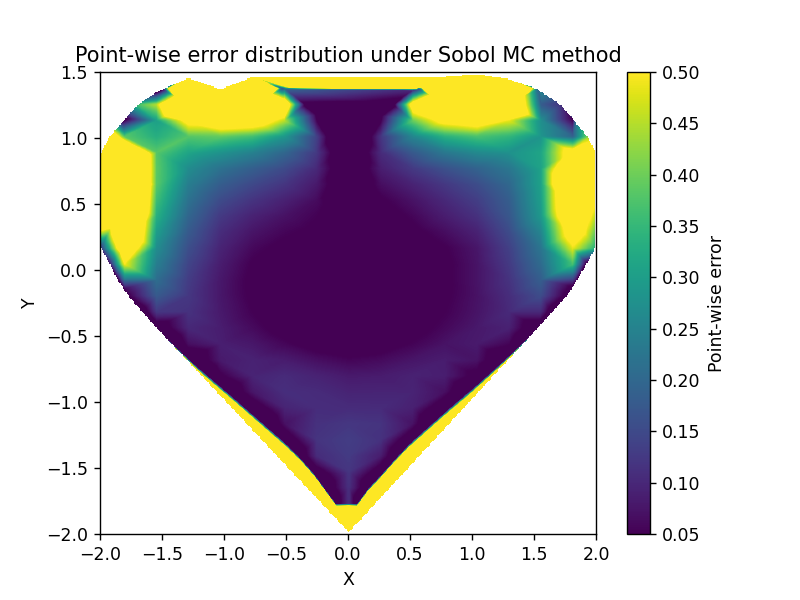}   
	\end{minipage}
}
\subfigure[] 
{
	\begin{minipage}[t]{0.4\linewidth}
	\includegraphics[width=1\textwidth]{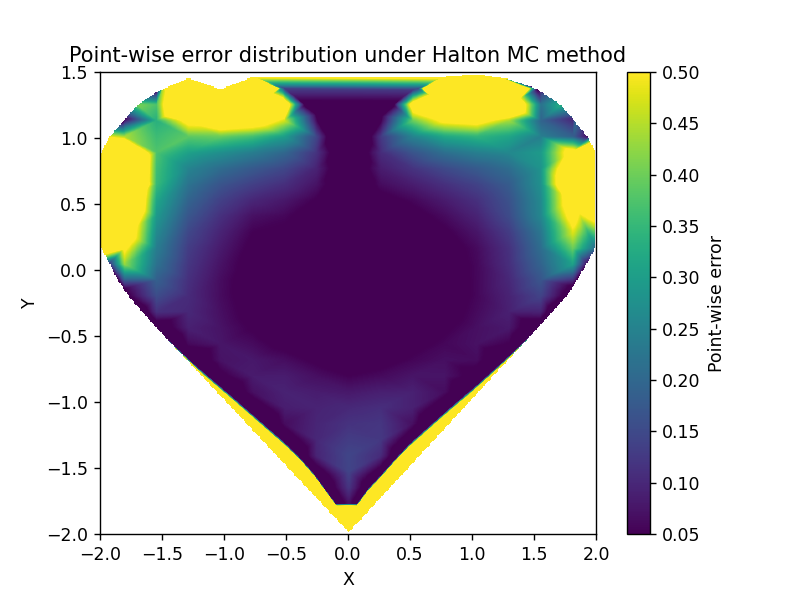}   
	\end{minipage}
}
\caption{Point-wise error distribution of Example 4.1 on $\Omega_2$ under different methods. (a) fPINN method, $L^2~\mathrm{error} = 9.43\mathrm{e}^{-2}$. (b) Monte Carlo method, $L^2~\mathrm{error} = 4.72\mathrm{e}^{-2}$. (c) Sobol Monte Carlo method, $L^2~\mathrm{error} = 4.19\mathrm{e}^{-2}$. (d) Halton Monte Carlo method, $L^2~\mathrm{error} = 4.51\mathrm{e}^{-2}$. Here we set $N=16$ and the network structure to 1 layer with 200 neurons.} 
\label{fig4}  
\end{figure}

We then similarly present the validity of the method for the above equations over the region $\Omega_2$ in Fig. \ref{fig4}. Again, it can be seen that our method is able to achieve the same or even better numerical accuracy, proving the effectiveness of our method again.\\

\textbf{Example 4.2.} Next, we consider the following 2D time-space fractional diffusion equation with initial and boundary conditions on three different regions $\Omega_1$
\begin{equation}
    \begin{aligned}
        _0^C D_t^{\alpha} u(x,y,t) =& -K_1 (-\triangle)_{xx}^{\beta_1} u(x, y, t) - K_2 (-\triangle)_{yy}^{\beta_2} u(x, y, t) \\
        &+ f(x, y, t),~(x, y, t)\in\Omega_i \times(0,T],\\
    \end{aligned}
\end{equation}
\begin{equation}
    u(x, y, 0)=u_0(x, y),~(x, y)\in\Omega_i,~u(x, y, t)=0,~(x, y)\in\partial\Omega_i \times(0,T].
\end{equation}
We consider the exact solution $u(x, y, t) = (x-x_{lb})^3 (x_{ub} - x)^3 (y-y_{lb})^3 (y_{ub} - y)^3 \cos t$, and thus
\begin{equation}
    f(x,y,t) = f_1(x,y,t) + f_2(x,y,t) + f_3(x,y,t).
\end{equation}
\begin{equation}
    f_1(x, y, t) = (x-x_{lb})^3 (x_{ub} - x)^3 (y-y_{lb})^3 (y_{ub} - y)^3 \cos(t+\frac{\pi\alpha}{2}),
\end{equation}
\begin{equation}
    \begin{aligned}
        f_2(x, y, t) &= \frac{1}{2cos(\pi\alpha/2)}\{\frac{|x_{lb}+x_{ub}|^3 \Gamma(4)}{\Gamma(4-\alpha)}((x - x_{lb})^{3-\alpha} + (x_{ub} - x)^{3-\alpha}) \\
        &- \frac{3 |x_{lb}+x_{ub}|^2 \Gamma(5)}{\Gamma(5-\alpha)}((x - x_{lb})^{4-\alpha} + (x_{ub} - x)^{4-\alpha}) \\
        &+ \frac{3 |x_{lb}+x_{ub}| \Gamma(6)}{\Gamma(6-\alpha)}((x - x_{lb})^{5-\alpha} + (x_{ub} - x)^{5-\alpha}) \\
        &- \frac{3\Gamma(7)}{\Gamma(7-\alpha)}((x - x_{lb})^{6-\alpha} + (x_{ub} - x)^{6-\alpha})\} (y-y_{lb})^3 (y_{ub} - y)^3 \cos(t),
    \end{aligned}
\end{equation}
\begin{equation}
    \begin{aligned}
        f_3(x,y,t) &= \frac{1}{2cos(\pi\beta/2)}\{\frac{|y_{lb}+y_{ub}|^3 \Gamma(4)}{\Gamma(4-\beta)}((y - y_{lb})^{3-\beta} + (y_{ub} - y)^{3-\beta}) \\
        &- \frac{3 |y_{lb}+y_{ub}|^2 \Gamma(5)}{\Gamma(5-\beta)}((y - y_{lb})^{4-\beta} + (y_{ub} - y)^{4-\beta}) \\
        &+ \frac{3 |y_{lb}+y_{ub}| \Gamma(6)}{\Gamma(6-\beta)}((y - y_{lb})^{5-\beta} + (y_{ub} - y)^{5-\beta}) \\
        &- \frac{3\Gamma(7)}{\Gamma(7-\beta)}((y - y_{lb})^{6-\beta} + (y_{ub} - y)^{6-\beta})\} (x-x_{lb})^3 (x_{ub}-x)^3 \cos(t),
    \end{aligned}
\end{equation}
For simplicity, we take $K_1 = 1$, $K_2 = 1$ and $T=1$ in this example. As in Example 4.1, we also present the effect of different number of approximation points on numerical accuracy and computational speed under different methods in Table 3. The results show that our method still achieves good numerical accuracy, and the computational speed advantage over the fPINN method is more obvious as the number of approximation points increases.
\renewcommand{\arraystretch}{1.5} 
\begin{table}[htp]

  \centering
  \fontsize{4}{4}\selectfont
  \begin{threeparttable}
  \caption{Comparison of performance between fPINNs \cite{fpinn} and our method on Example 4.2 (2D time-space fractional diffusion equation) on $\Omega_1$ for $\alpha=0.3, \beta_1=1.7, \beta_2=1.5$. For each method, we set the number of iterations to 100K.}
  \label{tab:performance_comparison}
    \begin{tabular}{ccccccccccccc}
    \toprule
    \multirow{1}{*}{N=K}&
    \multicolumn{2}{c}{ fPINN}&\multicolumn{2}{c}{ Monte Carlo PINN}&\multicolumn{2}{c}{ Sobol Monte Carlo PINN}&\multicolumn{2}{c}{ Halton Monte Carlo PINN}\cr
    \cmidrule(lr){2-3} \cmidrule(lr){4-5} \cmidrule(lr){6-7} \cmidrule(lr){8-9}
    &accuracy&time to run ten iterations&accuracy&time to run ten iterations&accuracy&time to run ten iterations&accuracy&time to run ten iterations\cr
    \midrule
    20&5.79e-01&0.09&2.59e-01&0.05&1.72e-01&0.05&3.06e-01&0.05\cr
    40&8.98e-02&0.13&7.64e-02&0.07&6.61e-02&0.07&7.34e-02&0.07\cr
    80&4.11e-02&0.24&4.01e-02&0.15&3.82e-02&0.15&4.12e-02&0.15\cr
    \bottomrule
    \end{tabular}
    \end{threeparttable}
\end{table}

\textbf{Example 4.3.} Here we consider the following 3D time-space fractional Bloch-Torrey equation on the ball $\Omega={(x,y,z):\sqrt{x^2 + y^2 + z^2}<r=1/2}$ with homogeneous initial and boundary conditions \cite{yang2020}
\begin{equation}\label{examp4.3}
    \begin{aligned}
        &K_{\alpha} ^C D_t^{\alpha} u (x, y, z, t) = K_{\beta} R^{2\beta} u(x, y, z, t) + f(x, y, z, t), \\
        &u(x, y, z, 0) = 0,~(x, y, z)\in\Omega, \\
        &u(x, y, z, t) = 0,~(x, y, z, t)\in R^3 \ \Omega\times(0,T],
    \end{aligned}
\end{equation}
where
\begin{equation}
    R^{2\beta} := ^{Rz}D_x^{2\beta} + ^{Rz}D_y^{2\beta} + ^{Rz}D_z^{2\beta},
\end{equation}
where $^{Rz}D_x^{2\beta}$ is the Riesz fractional derivative operator \cite{yang2020}
\begin{equation}
    ^{Rz}D_x^{2\beta} f(x) := -\frac{1}{2\cos(\pi\beta)}(_{x_{lb}}D_x^{2\beta} f(x) + _x D_{x_{ub}}^{2\beta} f(x)).
\end{equation}
With exact solution $u(x, y, z, t) = 2^4 t^2 (x^2 + y^2 + z^2 - r^2)^2$ and
\begin{equation}
    \begin{aligned}
        f(x, y, z, t) = &h(x,\sqrt{r^2 - y^2 - z^2}) + h(y,\sqrt{r^2 - x^2 - z^2}) + h(z,\sqrt{r^2 - x^2 - y^2}) \\
        &+ \frac{32K_\alpha}{\Gamma(3-\alpha)}t^{2-\alpha} (x^2 + y^2 + z^2 - r^2)^2
    \end{aligned}
\end{equation}
with
\begin{equation}
    \begin{aligned}
        h(x,s) = \frac{8K_\beta t^2}{cos(\beta\pi/2)}&(\frac{8s^2}{\Gamma(3-2\beta)}((x+s)^{2-2\beta} + (s-x)^{2-2\beta}) \\
        &- \frac{24s}{\Gamma(4-2\beta)}((x+s)^{3-2\beta} + (s-x)^{3-2\beta}) \\
        &+ \frac{24}{\Gamma(5-2\beta)}((x+s)^{4-2\beta} + (s-x)^{4-2\beta})).
    \end{aligned}
\end{equation}

\begin{figure}[htb]
\centering
\subfigure[] 
{
	\begin{minipage}[t]{0.4\linewidth}
	\includegraphics[width=1\textwidth]{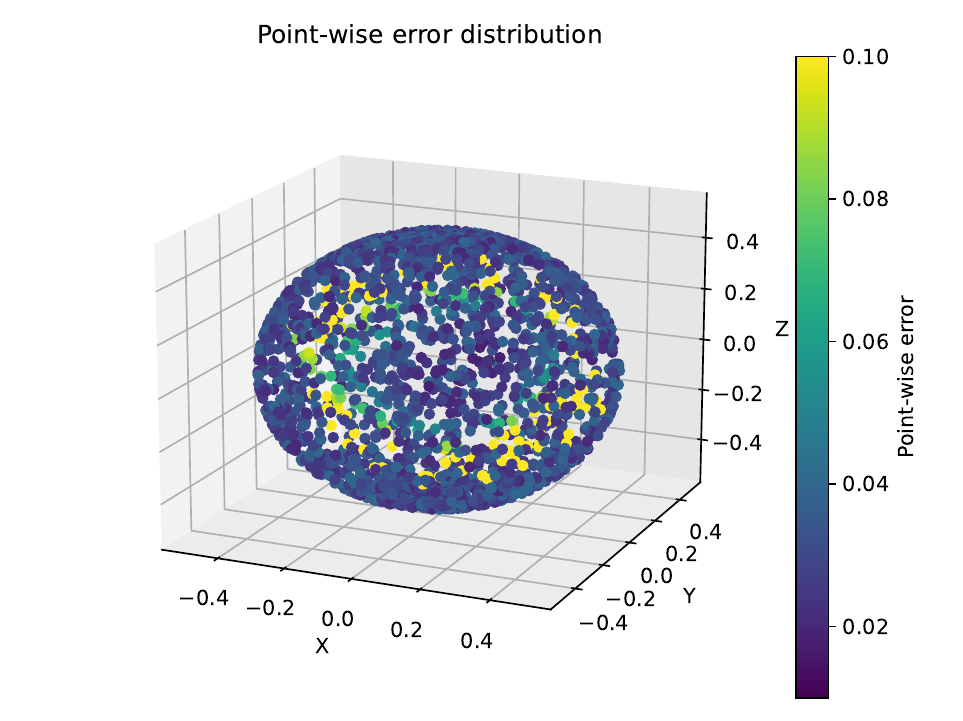}   
	\end{minipage}
}
\subfigure[] 
{
	\begin{minipage}[t]{0.4\linewidth}
	\includegraphics[width=1\textwidth]{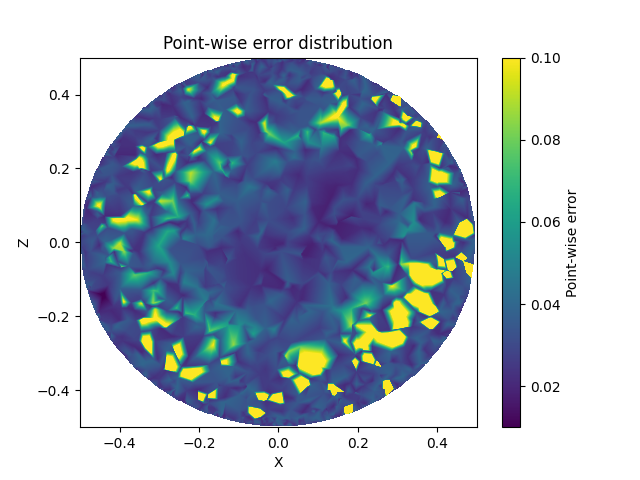}   
	\end{minipage}
}
\caption{Point-wise error distribution of Example 4.3 under Halton Monte Carlo method, $L^2~\mathrm{error} = 3.18\mathrm{e}^{-2}$. (a) Error distribution over the entire region. (b) Error distribution over the $y = 0$ cross section.} 
\label{fig5}  
\end{figure}

For simplicity, we take $K_\alpha = 1$, $K_\beta = 1$ and $T = 1$. Here, we only discuss the performance of the Halton Monte Carlo method for fitting the numerical solution to this problem. In Fig. \ref{fig5}, the overall point-wise error and the point-wise error on the $y = 0$ cross section are shown in Fig. \ref{fig5}(a) and Fig. \ref{fig5}(b), respectively. As can be seen from the figures, our method is able to solve the 3D fractional irregular region problem and achieve good numerical accuracy, again proving the effectiveness of our method.

\begin{figure}[htb]
\centering
\subfigure[] 
{
	\begin{minipage}[t]{0.3\linewidth}
	\includegraphics[width=1\textwidth]{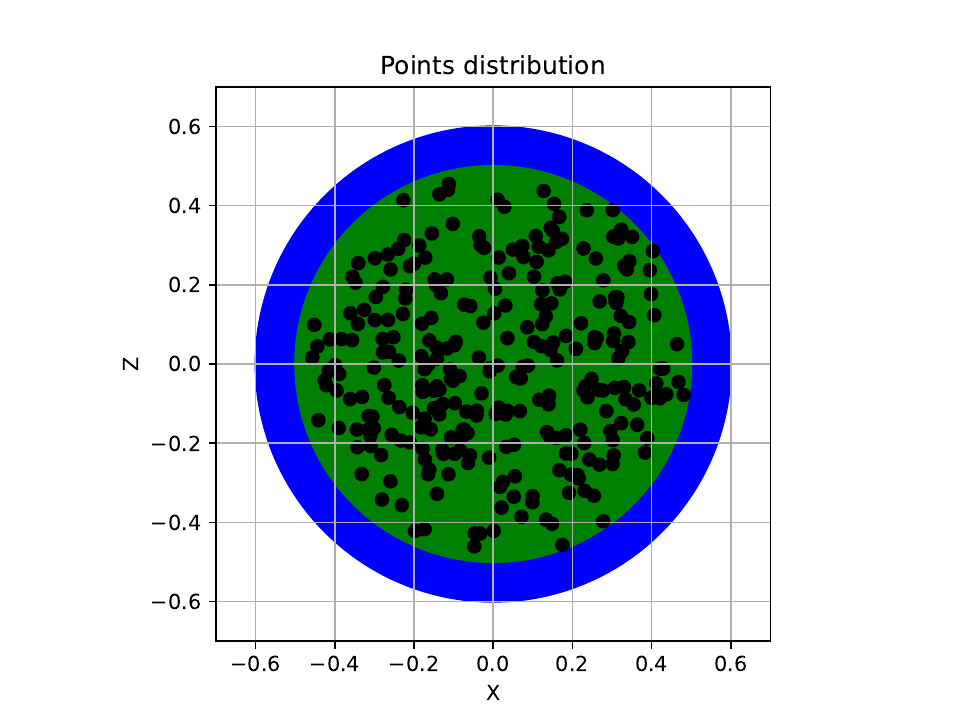}   
	\end{minipage}
}
\subfigure[] 
{
	\begin{minipage}[t]{0.3\linewidth}
	\includegraphics[width=1\textwidth]{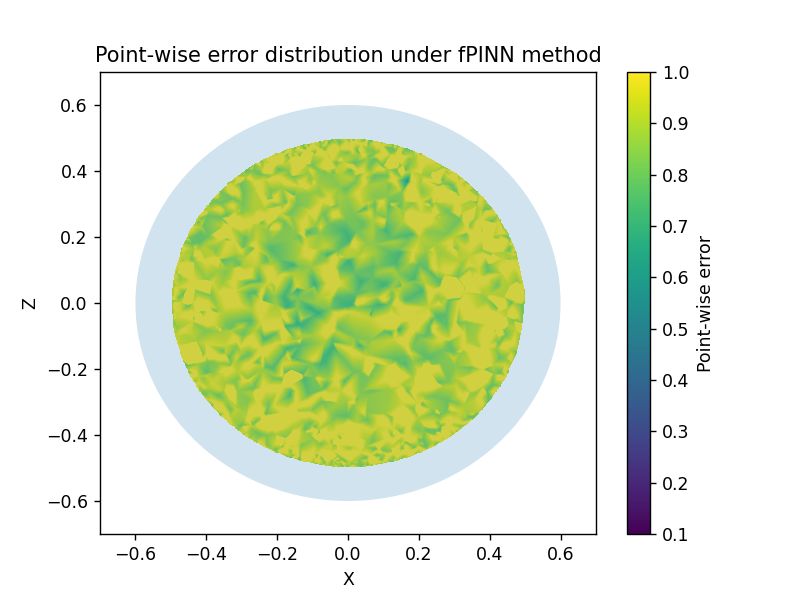}   
	\end{minipage}
}
\subfigure[] 
{
	\begin{minipage}[t]{0.3\linewidth}
	\includegraphics[width=1\textwidth]{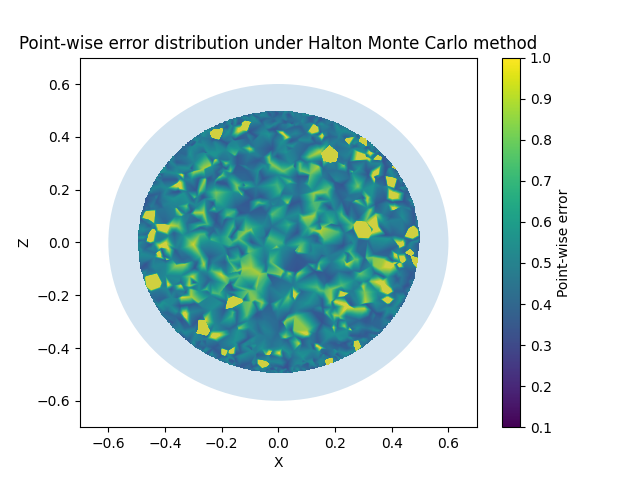}   
	\end{minipage}
}
\caption{Point-wise error distribution of fuzzy boundary location problem under different methods. (a) dataset. (b) fPINN method, $L^2~\mathrm{error} = 9.18\mathrm{e}^{-1}$. (c) Halton Monte Carlo method, $L^2~\mathrm{error} = 4.17\mathrm{e}^{-1}$.} 
\label{fig6}  
\end{figure}

Then, we test a very interesting and practical problem. Consider Example 4.3 where the location of the boundary is fuzzy, i.e., the location of its boundary is not known exactly. For simplicity, assume that the real boundary is outside the wall of a sphere of radius 0.5, but the actual measured fuzzy boundary is the wall of a sphere of radius 0.6. We performed numerical simulation of this model and comparison of the numerical results of our method with the original fPINN simulation of it. We show the results in Fig. \ref{fig6}. From the figure we can see that our method is able to achieve better numerical accuracy compared to the original fPINN method.

\textbf{Example 4.4.} Finally, we consider the following 3D coupled time-space fractional Bloch-Torrey equation \cite{yang2020}
\begin{equation}
    \begin{aligned}
        &\omega^{\alpha-1} {}^C D_t^{\alpha} M_x (x, y, z, t) = D\mu^{2\beta - 2}R^{2\beta}M_x(x, y, z, t) + \lambda(t)M_y(x, y, z, t),\\
        &\omega^{\alpha-1} {}^C D_t^{\alpha} M_y (x, y, z, t) = D\mu^{2\beta - 2}R^{2\beta}M_y(x, y, z, t) + \lambda(t)M_x(x, y, z, t),
    \end{aligned}
\end{equation}
with
\begin{equation}
    \begin{aligned}
        &M_x(x, y, z, 0) = 0,~M_y(x, y, z, 0) = 100,~\forall (x, y, z)\in\Omega,\\
        &M_x(x, y, z, t) = 0,~M_y(x, y, z, t) = 0,~\forall (x, y, z, t)\in \mathbb{R}^3\backslash\Omega\times(0,T],
    \end{aligned}
\end{equation}
\begin{figure}[htbp]
\centering
\includegraphics[width=0.7\linewidth]{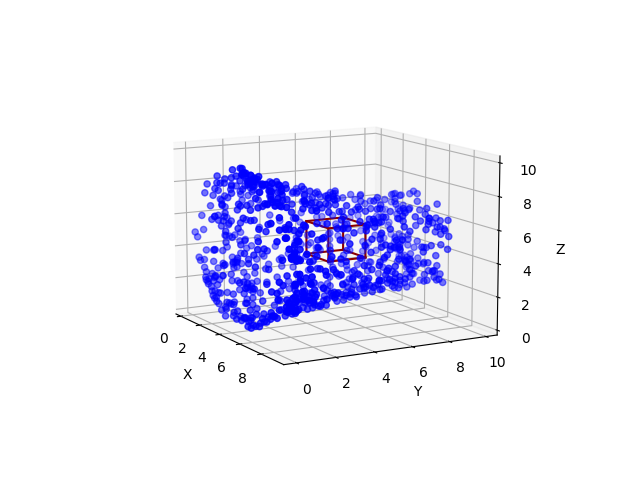}\\
\caption{Residual points for Example 4.4 (length units in mm).}
\label{fig7}
\end{figure}
where $\lambda(t) = z\gamma G_z$ and denote the constant $g = \gamma G_z = 100 \text{mm}^{-1} \text{s}^{-1}$. It is well known that since the diffusion of water molecules can be different for different tissue structures, the apparent diffusion coefficient (ADC) contrast is different for each tissue. The value of ADC usually determined by fitting the experimental data with the solution of the Bloch-Torrey equation. In this example, the magnetization is defined as $|M_{xy}| = \sqrt{M_x^2 + M_y^2}$, which is the signal we receive in the MRI device. For more information about the model, the reader is referred to the literature \cite{magin2008}.

In this example, we set $T=100~\text{ms}$ and the diffusion coefficient $D$ is $1\times 10^{-3} \text{mm}^2 / \text{s}$; also, $\omega = 2\times 10^{-3} \text{s}$ and $\mu = 5\times 10^{-3} \text{mm}$.

\begin{figure}[htb]
\centering
\subfigure[] 
{
	\begin{minipage}[t]{0.3\linewidth}
	\includegraphics[width=1\textwidth]{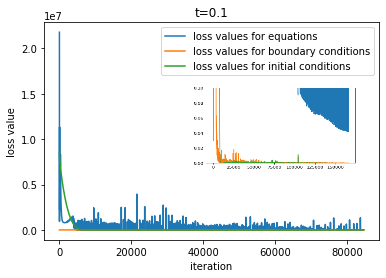}   
	\end{minipage}
}
\subfigure[] 
{
	\begin{minipage}[t]{0.3\linewidth}
	\includegraphics[width=1\textwidth]{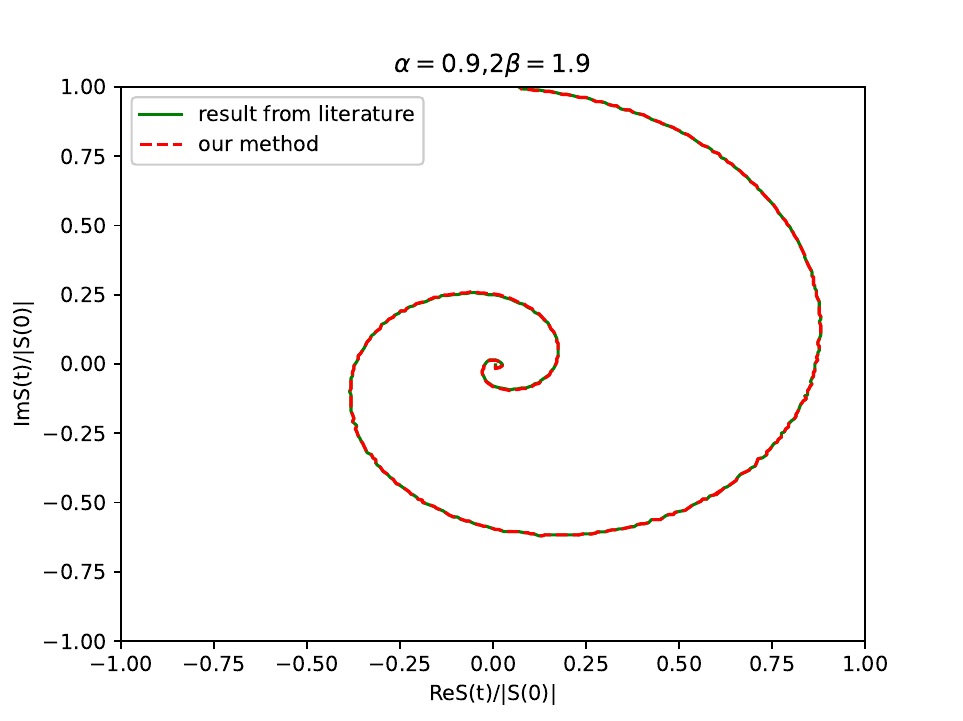}   
	\end{minipage}
}
\subfigure[] 
{
	\begin{minipage}[t]{0.3\linewidth}
	\includegraphics[width=1\textwidth]{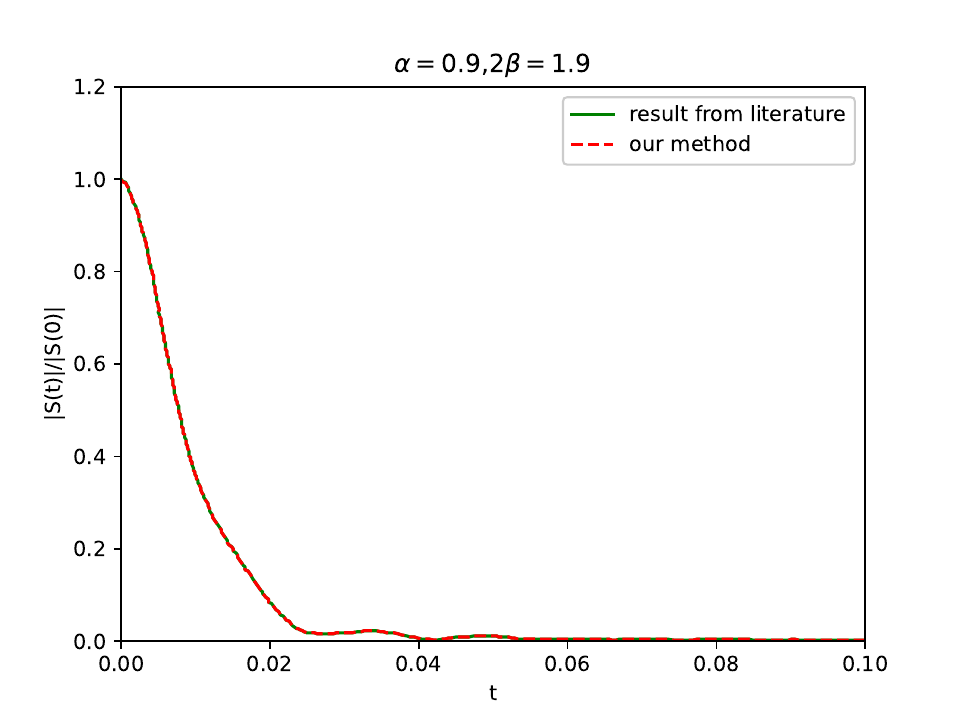}   
	\end{minipage}
}
\caption{Simulation results for Example 4.4 under $\alpha=0.9, 2\beta=1.9$. (a) Loss values at $t=0.1$. Comparison with results in literature \cite{yang2020}: (b) Plot of $\mathrm{Re}|S(t)|/|S(0)|$ versus $\mathrm{Im}|S(t)|/|S(0)|$ of $\Omega_{\mathrm{ROI}}$ from $t=0$ to $t=0.1$. (c) Plot of variation of $|S(t)|/|S(0)|$ of the region $\Omega_{\mathrm{ROI}}$.} 
\label{fig8}  
\end{figure}
We constructed the dataset using only 1000 model interior points, as shown in Fig \ref{fig7}. The signal is defined by
\begin{equation}
    S(t) = \int_{\Omega_{\mathrm{ROI}}} M_{xy}(\mathbf{x}, t)d\mathbf{x},
\end{equation}
and
\begin{equation}
    \mathrm{Re}S(t) = \int_{\Omega_{\mathrm{ROI}}} M_{x}(\mathbf{x}, t)d\mathbf{x},~\mathrm{Im}S(t) = \int_{\Omega_{\mathrm{ROI}}} M_{y}(\mathbf{x}, t)d\mathbf{x},
\end{equation}
where $\Omega_{\mathrm{ROI}}\in\Omega$ is a cube with length 2 mm in the middle of $\Omega$, as shown the square in Fig. \ref{fig7}. Since the model is more complex and taking more training points will lead to slower computation, the 4D data space covered by the model will be very sparse if the temporal variable $t$ and spatial variables $x,y,z$ are randomly sampled and paired together directly under the premise of taking 1000 pairs of equations for training points. Therefore, we use here an iterative optimization training method, in which we train time $t$ sequentially from small to large, and ensure that the coordinates of the $x, y, z$ data sets are the same at each time layer. We show the results in Fig. \ref{fig8}, from which we can see that the numerical results are in good agreement with those obtained by the traditional numerical methods.

\section{Summary}\label{sec5}
In this paper, we developed a new (quasi) Monte Carlo method for approximating fractional differentiation  and successfully applied it to solving fPDEs by combining it with PINNs. This method is not only able to approximate the Gr$\ddot{\mathrm{u}}$nwald-Letnikov, Caputo, and Riemann-Liouville fractional derivatives, but also due to the lumped distribution nature of the sample dataset generated by the Monte Carlo method, it is computationally fast while ensuring numerical accuracy. Several numerical examples demonstrated the effectiveness of our method, including a realistic application. By comparing our method with the original fPINN method \cite{fpinn} and the Monte Carlo fPINN method \cite{lingguo}, we found that our method has better numerical accuracy as well as faster computation speed, and its applicability is broader compared to the Monte Carlo fPINN method. Moreover, our approach allows parallelization of computations on GPUs, although this was not pursued in the current study.

\section*{Declaration of competing interest}
Shupeng Wang declares that he has no known competing financial interests or personal relationships that could have appeared to influence the work reported in this paper.
George Karniadakis declares that he has financial interests with the companies Anailytica and PredictiveIQ.

\section*{Data availability}
Data will be made available on request.

\section*{Acknowledgments}
The authors would like to thank Prof. Ling Guo for sharing code during the preparation of this work.

\end{document}